\documentclass[10pt,twocolumn,letterpaper]{article}

\usepackage{iccv}
\usepackage{times}
\usepackage{epsfig}
\usepackage{graphicx}
\usepackage{amsmath}
\usepackage{amssymb}

\usepackage{subfigure}
\usepackage{multirow}
\usepackage{bbding}

\newtheorem{theorem}{Theorem}

\newtheorem{remark}{Remark}

\newtheorem{corollary}{Corollary}

\newcommand{\name}{Cyclic Output Discrepancy}
\newcommand{\abbr}{COD}

\usepackage{xcolor}
\definecolor{ultramarine}{RGB}{18, 10, 143} 
\definecolor{xgreen}{RGB}{0,192,0} 
\usepackage[pagebackref=true,breaklinks=true,letterpaper=true,colorlinks,bookmarks=false,citecolor=xgreen]{hyperref}
\iccvfinalcopy % *** Uncomment this line for the final submission

\def\iccvPaperID{1177} % *** Enter the ICCV Paper ID here

\ificcvfinal\fi

\begin{document}

%%%%%%%%% TITLE
\title{Semi-Supervised Active Learning with Temporal Output Discrepancy}

\author{Siyu Huang$^{1}$\quad\quad
Tianyang Wang$^{2}$ \quad\quad
Haoyi Xiong$^1$ \quad\quad
Jun Huan$^3$ \quad\quad
Dejing Dou$^1$ \\
$^1$Baidu Research \quad\quad\quad
$^2$Austin Peay State University \quad\quad\quad
$^3$Styling AI \\
{\tt\small \{huangsiyu,xionghaoyi,doudejing\}@baidu.com} \quad
{\tt\small wangt@apsu.edu} \quad
{\tt\small lukehuan@shenshangtech.com} \\
}

% \author{Siyu Huang\\
% Baidu Research\\
% {\tt\small huangsiyu@baidu.com}
% \and
% Tianyang Wang\\
% Austin Peay State University\\
% {\tt\small toseattle@siu.edu}
% \and
% Haoyi Xiong\\
% Baidu Research\\
% {\tt\small xionghaoyi@baidu.com}
% \and
% Jun Huan\\
% Styling AI\\
% {\tt\small lukehuan@shenshangtech.com}
% \and
% Dejing Dou\\
% Baidu Research\\
% {\tt\small doudejing@baidu.com}
% }
\maketitle
\ificcvfinal\thispagestyle{empty}\fi

%%%%%%%%% ABSTRACT
\begin{abstract}
While deep learning succeeds in a wide range of tasks, it highly depends on the massive collection of annotated data which is expensive and time-consuming. To lower the cost of data annotation, active learning has been proposed to interactively query an oracle to annotate a small proportion of informative samples in an unlabeled dataset. Inspired by the fact that the samples with higher loss are usually more informative to the model than the samples with lower loss, in this paper we present a novel deep active learning approach that queries the oracle for data annotation when the unlabeled sample is believed to incorporate high loss. The core of our approach is a measurement Temporal Output Discrepancy (TOD) that estimates the sample loss by evaluating the discrepancy of outputs given by models at different optimization steps. Our theoretical investigation shows that TOD lower-bounds the accumulated sample loss thus it can be used to select informative unlabeled samples. On basis of TOD, we further develop an effective unlabeled data sampling strategy as well as an unsupervised learning criterion that enhances model performance by incorporating the unlabeled data. Due to the simplicity of TOD, our active learning approach is efficient, flexible, and task-agnostic. Extensive experimental results demonstrate that our approach achieves superior performances than the state-of-the-art active learning methods on image classification and semantic segmentation tasks. 
\end{abstract}

%%%%%%%%% BODY TEXT
\section{Introduction}
Large-scale annotated datasets are indispensable and critical to the success of modern deep learning models. Since the annotated data is often highly expensive to obtain, learning techniques including unsupervised learning \cite{unsupervised2015}, semi-supervised learning \cite{semi2007}, and weakly supervised learning \cite{weak2010} have been widely explored to alleviate the dilemma. In this paper\footnote{Code is available at \url{https://github.com/siyuhuang/TOD}} we focus on active learning \cite{active1996} which aims to selectively annotate unlabeled data with limited budgets while resulting in high performance models. 

In existing literature of active learning, two mainstream approaches have been studied, namely the diversity-aware approach and the uncertainty-aware approach. The diversity-aware approach \cite{diverse2010} aims to pick out diverse samples to represent the distribution of a dataset. It works well on low-dimensional data and classifier with a small number of classes \cite{coreset}. The uncertainty-aware approach \cite{erm2015} aims to pick out the most uncertain samples based on the current model. However, the uncertainty heuristics, such as distance to decision boundary \cite{boundary2003} and entropy of posterior probabilities \cite{entropy2008}, are often task-specific and need to be specifically designed for individual tasks such as image classification \cite{entropy2009}, object detection \cite{detection}, and semantic segmentation \cite{segmentation}.

%In this paper, we present a simple and task-agnostic active learning approach for deep neural networks. We are inspired by the evidences as follow. 
In this paper, we consider that the samples with higher loss would be more informative than the ones with lower loss. Specifically in supervised learning settings, when samples are correctly labeled, the averaged loss function over all samples should be gradually minimized during the learning procedure. Moreover, in every iteration the training model would backward propagated error according to the loss of every sample~\cite{lecun1988theoretical}, while the sample with high loss usually brings informative updates to the parameters of the training model~\cite{he2020local}. In this work, we generalize these evidences to active learning problems and propose a simple yet effective loss estimator Temporal Output Discrepancy~(TOD), which could measure the potential loss of a sample only relied on the training model, when the ground-truth label of the sample is not available. Specifically, TOD computes the discrepancy of outputs given by models at different optimization steps, and a higher discrepancy corresponds to a higher sample loss. Our theoretical investigation shows that TOD~well measures the sample loss.
%optimized using gradient descent. 

%In this way, for every iteration of deep learning, the sample with higher loss would bring the training more information. 
%Our approach is inspired by the fact that the deep learning models are learned to minimize the loss function. The loss function can be regarded as a good metric for evaluating the amount of information, \ie, a sample with a higher loss is usually more informative to the model than a sample with a lower loss. To accurately estimate the loss of unlabeled samples, we propose a novel measurement \name~(\abbr). \abbr~computes the discrepancy between model outputs of two active learning cycles and a higher discrepancy corresponds to a higher sample loss. We theoretically show that \abbr~approximates the sample loss when a model is optimized using gradient descent. 

On basis of TOD, we propose a deep active learning framework that leverages a novel unlabeled data sampling strategy for data annotation in conjunction with a semi-supervised training scheme to boost the task model performance with unlabeled data. Specifically, the active learning procedure can be split into a sequence of training cycles starting with a small number of labeled samples. By the end of every training cycle, our data sampling strategy estimates \name~(\abbr), which is a variant of TOD, for every sample in the unlabeled pool and selects the unlabeled samples with the largest \abbr~for data annotation. The newly-annotated samples are added to the labeled pool for model training in the next cycles. Furthermore, with the aid of the unlabeled samples, we augment the task learning objective with a regularization term derived from TOD, so as to improve the performance of active learning in a semi-supervised manner.
%After every active learning cycle, the proposed data sampling strategy selects the samples with the largest \abbr~from the unlabeled pool and annotate them using human oracles. 
%The newly annotated samples are then added to the labeled pool for the next active learning cycle. On the other hand, since the temporal output discrepancy is an effective loss estimation method, it can be adopted as an unsupervised criterion to improve task model learning with unlabeled data. In a semi-supervised learning framework, we train the model with both a task loss on labeled data and a temporal consistency regularization on unlabeled data, leading to a better task performance.  

%Incorporating \abbr, we develop an unlabeled data sampling strategy and a semi-supervised training scheme for active learning. After every active learning cycle, the proposed data sampling strategy selects the samples with the largest \abbr~from the unlabeled pool and annotate them using human oracles. The newly annotated samples are then added to the labeled pool for the next active learning cycle. On the other hand, since the temporal output discrepancy is an effective loss estimation method, it can be adopted as an unsupervised criterion to improve task model learning with unlabeled data. In a semi-supervised learning framework, we train the model with both a task loss on labeled data and a temporal consistency regularization on unlabeled data, leading to a better task performance.  

Compared with the existing deep active learning algorithms, our approach is more efficient, more flexible, and easier to implement, since it does not introduce extra learnable models such as the loss prediction module \cite{ll4al} or the adversarial network \cite{vaal,sraal} for uncertainty estimation. In the experiments, our active learning approach shows superior performances in comparison with the state-of-the-art baselines on various image classification and semantic segmentation datasets. Further ablation studies demonstrate that our proposed TOD can well estimate the sample loss and benefit both the active data sampling and the task model learning. 

The contributions of this paper are summarized as follows.
\begin{enumerate}
    \item This paper proposes a simple yet effective loss measure TOD. Both theoretical and empirical studies validate the efficacy of TOD.
    \item This paper presents a novel deep active learning framework by incorporating TOD~into an active sampling strategy and a semi-supervised learning scheme.
    \item Extensive active learning experiments on image classification and semantic segmentation tasks evaluate the effectiveness of the proposed methods.
\end{enumerate}

%A typical approach is to discriminate the labeling states, \ie labeled or unlabeled, of samples using a discriminator, for instance an adversarial network \cite{vaal,sraal} used in recent works.
%  and it does not take into consideration the states of models

\section{Related Work}
%In this section, we introduce the preliminary studies of our work, where the most relevant studies would be discussed. 
\noindent
\textbf{Active Learning.}
Active learning aims to incrementally annotate samples that result in high model performance and low annotation cost \cite{active1996}. Active learning has been studied for decades of years and the existing methods can be generally grouped into two categories: the query-synthesizing approach and the query-acquiring approach. The query-synthesizing approach \cite{synthesize2017,synthesize2018} employs generative models to synthesize new informative samples. For instance, ASAL \cite{synthesize2020} uses generative adversarial networks (GANs) \cite{gan} to generate high-entropy samples. In this paper, we focus on the query-acquiring active learning which develops effective data sampling strategies to pick out the most informative samples from the unlabeled data pool.

The query-acquiring methods can be categorized as diversity-aware and uncertainty-aware methods. The diversity-aware methods \cite{diverse2004,diverse2010} select a set of diverse samples that best represents the dataset distribution. A typical diversity-aware method is the core-set selection \cite{coreset} based on the core-set distance of intermediate features. It is theoretically and empirically proven to work well with a small scale of classes and data dimensions. 

The uncertainty-aware methods \cite{kapoor2007active,erm2015,drop2016,drop2017,bayesian2019} actively select the most uncertain samples in the context of the training model. A wide variety of related methods has been proposed, such as Monte Carlo estimation of expected error reduction \cite{montecarlo2001}, distance to the decision boundary \cite{boundary2001,boundary2003}, margin between posterior probabilities \cite{margin2006}, and entropy of posterior probabilities \cite{entropy2008,entropy2009,entropy2013}. 

The diversity-aware and uncertainty-aware approaches are complementary to each other thus many hybrid methods \cite{hybrid2013,hybrid2016,hybrid2017a,hybrid2017b,hybrid2017c,hybrid2018} have been proposed for specific tasks. In more recent literature, adversarial active learning \cite{dfal,vaal,sraal} is introduced to learn an adversarial discriminator to distinguish the labeled and unlabeled data.

Compared to the existing works in active learning, our method falls into the category of uncertainty-aware active learning by directly utilizing the task model for uncertainty estimation. The relevant works include the ones which utilize the expected gradient length \cite{gradient2008} or output changes on input perturbation \cite{outputchange2014,outputchange2016} for uncertainty estimation. In the realm of loss estimation, Yoo \etal \cite{ll4al} propose to learn a loss prediction module to estimate the loss of unlabeled samples. Different from existing methods which require extra deep models such as loss prediction network \cite{ll4al} or adversarial network \cite{vaal,sraal} for uncertainty estimation, we propose a learning-free principle for efficient active learning by evaluating the discrepancy of model outputs at different active learning cycles. Except its efficiency and task-agnostic property, we demonstrate that it is a lower bound of the accumulated sample loss, ensuring that the data samples of loss of higher lower bound can be picked out.

\noindent
\textbf{Semi-Supervised Learning.}
This work is also related to semi-supervised learning which seeks to learn from both labeled and unlabeled data, since we also develop the proposed loss estimation method to improve the learning of task model using unlabeled data. There has been a wide variety of semi-supervised learning approaches such as transductive model \cite{semitransductive2003}, graph-based method \cite{semigraph}, and generative model \cite{semigen2014}. We refer to \cite{semisurvey} for an up-to-date overview. 

%More related to this work, 
More recently, several semi-supervised methods including $\Pi$-model \cite{temporalensembling} and Virtual Adversarial Training \cite{vat} apply consistency regularization to the posterior distributions of perturbed inputs. Further improvements including Mean Teacher \cite{emanet} and Temporal Ensembling \cite{temporalensembling} apply the consistency regularization on models at different time steps. However, the consistency regularization has been seldom exploited for active learning.

%In this paper, we develop temporal consistency regularization into active learning
% Different from the existing methods, TOD~is naturally adapted to active learning since it is only calculated based on the model state of the end of the last active learning cycle.

%In addition, d
Compared to the existing efforts in semi-supervised learning for neural networks, our proposed loss measure TOD could be considered as an alternative solution of consistency regularization. TOD can be well adapted to active learning by developing a novel active sampling method COD. COD only relies on the models learned after every active learning cycle. In contrast, the existing temporal consistency-based uncertainty measurements often require access to a number of previous model states. For instance, the computing of Mean Teacher \cite{emanet} and Temporal Ensembling \cite{temporalensembling} require the historical model parameters and the historical model outputs, respectively. 

On the other hand, there have not been sufficient theoretical interpretations for the success of consistency regularization. Athiwaratkun \etal \cite{swa} reveals that the consistency regularization on perturbed inputs is an unbiased estimator for the norm of the Jacobian of the network. However, there is still a lack of interpretations on the temporal consistency regularization. In this paper, we show that the temporal consistency regularization can be connected to the lower bound of the accumulated sample loss. Thus, the temporal consistency regularization is an theoretically effective solution to loss estimation as well as semi-supervised learning. 

\section{Temporal Output Discrepancy} \label{sec:cod}

%A high TOD value indicates a high uncertainty of the sample.
%
%Data sampling strategy is at the core of active learning. A successful sampler is expected to pick out the most informative/uncertain samples from the unlabeled pool conditioned on a task model $f$. It is a challenging task to develop the sampling strategy since the labels cannot be observed by the sampler. 
%In this work, 
 %Here we introduce a simple measurement,  to effectively and efficiently select the most informative samples from the unlabeled pool. 

Measuring the sample loss on a given neural network $f$, when the label of the sample is unavailable, is a key challenge for many learning problems, including active learning \cite{margin2006,entropy2013, drop2017}, continual learning \cite{bayesian2019}, and self-supervised learning \cite{emanet, temporalensembling}. In this work, we present Temporal Output Discrepancy~(TOD), which estimates the sample loss based on the discrepancy of outputs of a neural network at different learning iterations. Given a sample $x \in \mathbb{R}^d$, we have TOD $D_t^{\{T\}}:\mathbb{R}^d \rightarrow \mathbb{R}$
\begin{equation}
    D_t^{\{T\}}(x) \overset{\text{def}}{=} \|f(x;w_{t+T}) - f(x;w_{t})\| .
\end{equation}
$D_t^{\{T\}}(x)$ characterizes the distance\footnote{For brevity, $\|\cdot\|$ denotes the $L_2$ norm $\|\cdot\|_2$ in this paper.} between outputs of model $f$ with parameters $w_{t+T}$ and $w_{t}$ obtained in the $(t+T)$-th and $t$-th gradient descend step during learning (\eg, $T>0$), respectively. 
%$ D_t^{\{T\}}(x)$ denotes the output discrepancy brought by $T$-step gradient descent. 

\begin{figure}[t]
    \centering
    \includegraphics[width=0.495\linewidth]{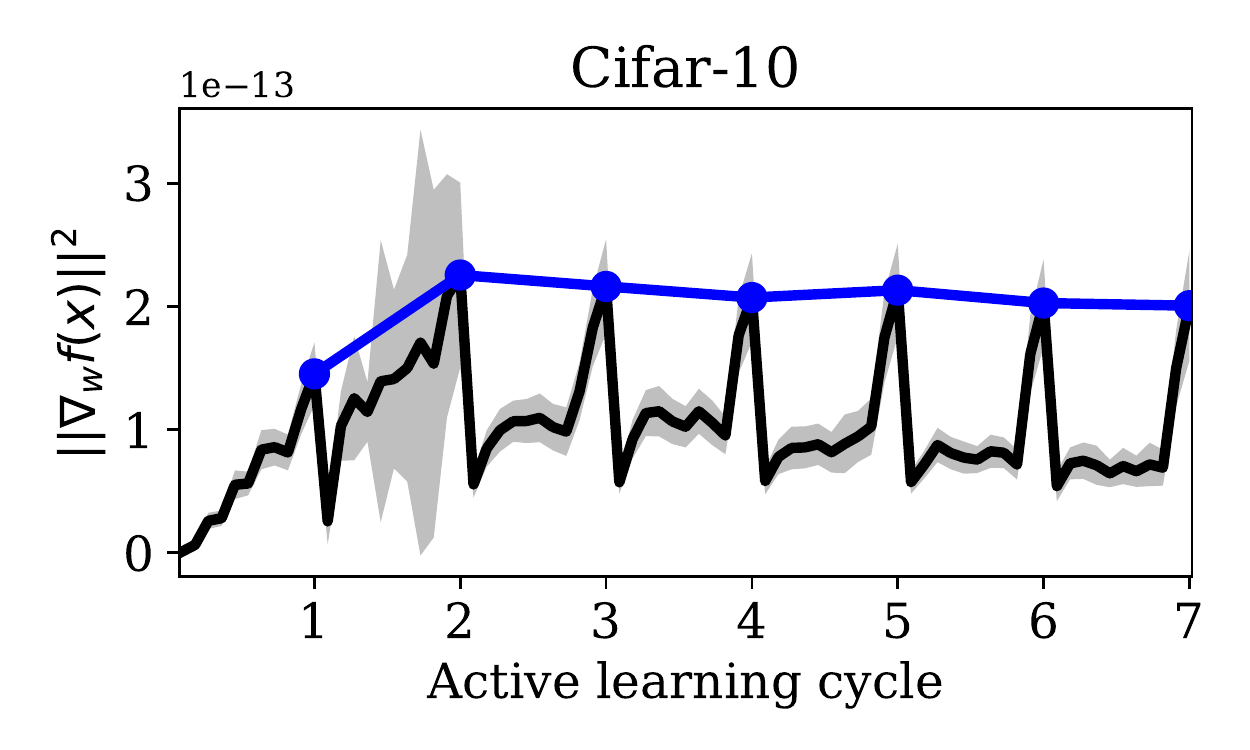}
    \includegraphics[width=0.495\linewidth]{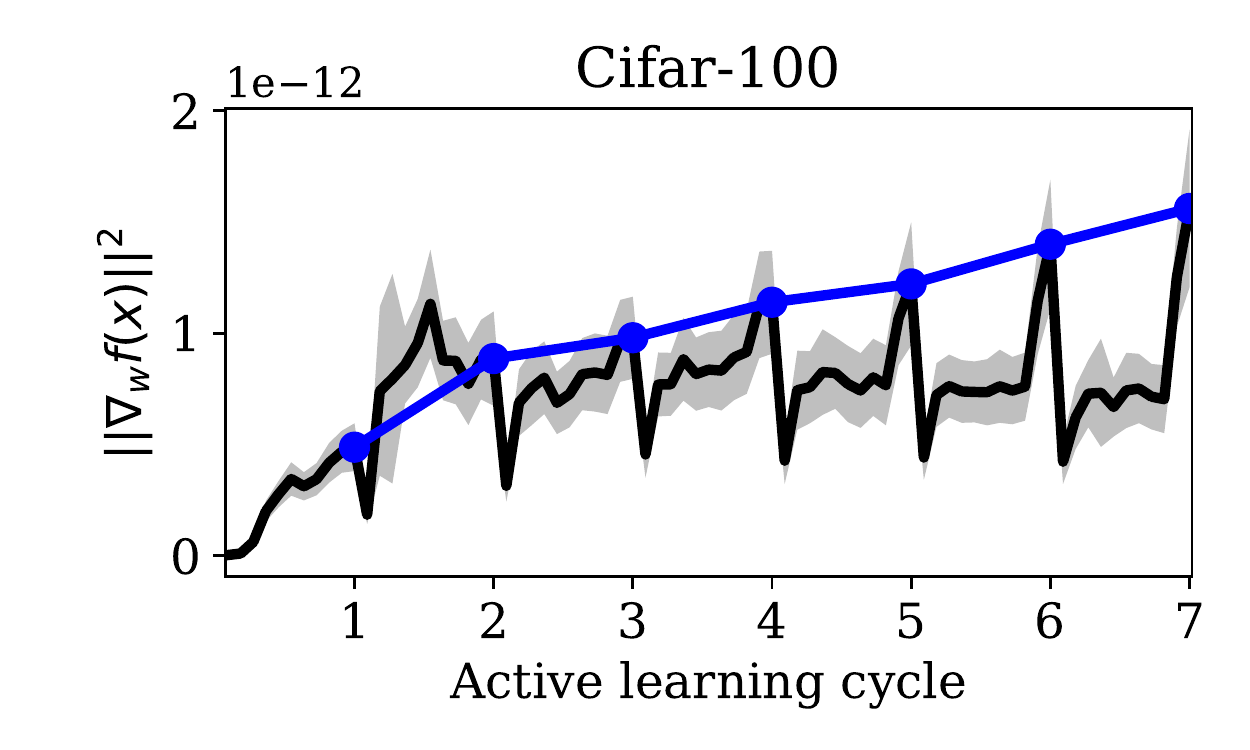}
    \caption{$\| \nabla_w f \|^2$, \vs, the active learning cycle. The dark lines denote $\| \nabla_w f \|^2$ averaged over the training (\ie, labeled and unlabeled) pool. The blue lines denote average $\| \nabla_w f \|^2$ after every active learning cycle.}
    \label{fig:dW}
\end{figure}

In the following, we show that a larger $D_t^{\{T\}}(x)$ indicates a larger sample loss\footnote{Here we take Euclidean loss as an example. The cross-entropy loss has similar results.} $L_t(x) = \frac{1}{2}(y - f(x;w_t))^2$, where $y \in \mathcal{R}$ is the label corresponding to sample $x$. We first give the upper bound of one-step output discrepancy $D_t^{\{1\}}(x)$.
\begin{theorem} \label{theorem}
With an appropriate setting of learning rate $\eta$,
\begin{equation}
    \label{eq:theorem1}
     D_t^{\{1\}}(x) \leq \eta \sqrt{2L_t(x)} \| \nabla_{w} f(x;w_t)\|^2 . 
\end{equation}
\end{theorem}

\begin{figure*}[t]
    \centering
    \includegraphics[width=0.87\linewidth]{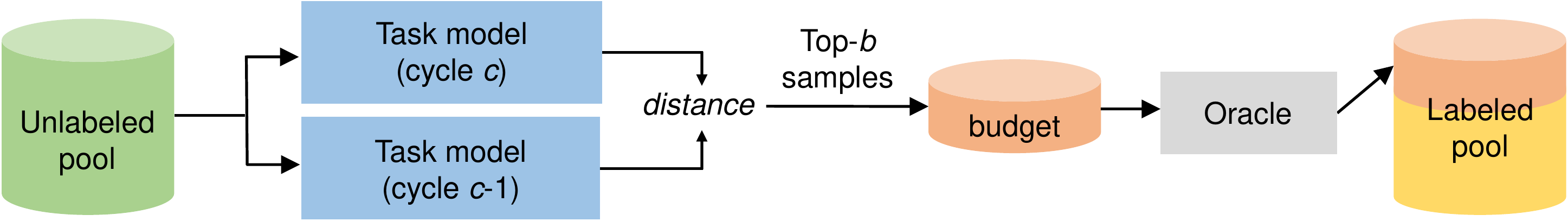}
    \vspace{.5em}
    \caption{The \abbr-based unlabeled data sampling strategy for active learning. Data samples with the largest \abbr~are collected from the unlabeled pool. The collected samples are annotated by an oracle and 
    added to the labeled pool.}
    \label{fig:sampler}
    \vspace{.0em}
\end{figure*}

The proofs of Theorem \ref{theorem} and the following corollaries can be found in the supplementary material. From Theorem \ref{theorem}, the upper bound of $T$-step output discrepancy $D_t^{\{T\}}(x)$ can be easily deduced.
\begin{corollary} \label{corollary1}
With an appropriate setting of learning rate $\eta$,
\begin{equation}
    \label{eq:T_step}
     D_t^{\{T\}}(x) \leq  \sqrt{2} \eta \sum_{\tau=t}^{t+T-1} \left( \sqrt{L_\tau(x)} \| \nabla_{w} f(x;w_\tau)\|^2 \right)  .
\end{equation}
\end{corollary}

Corollary \ref{corollary1} preliminarily connects $T$-step output discrepancy $D_t^{\{T\}}(x)$ to sample loss $L(x)$. However, it is almost infeasible to compute $\|\nabla_{w} f(x;w_\tau)\|$ on all the $\tau$. Fortunately, $\| \nabla_w f \|$ is approximately a constant under the context of neural networks, as discussed in \cite{lip2013,lip2018}.
\begin{remark}
For a linear layer $\phi(x;W)$ with ReLU activation, the Lipschitz constant $\mathcal{L}(W) \leq \|x\|$.
\end{remark}

Since sample $x$ is drawn from a distribution $X$, we assume $\|x\|$ is upper-bounded by a constant so that $f$ is Lipschitz-continuous over $w$. Thus, we let $\| \nabla_w f \|^2$ be upper-bounded by a constant $C$. Empirical results on image classification benchmarks including Cifar-10 and Cifar-100 also support this assumption. As shown in Fig. \ref{fig:dW}, the dark lines are the averaged $\| \nabla_w f \|^2$ over the training set. The blue lines denote the averaged $\| \nabla_w f \|^2$ after every active learning cycle. $\| \nabla_w f \|^2$ has a small variance over samples and it is nearly constant across every active learning cycle. 

With $\| \nabla_w f \|^2 \leq C$, we rewrite Corollary \ref{corollary1} to connect $D_t^{\{T\}}(x)$ with the accumulated loss of sample $x$.
\begin{corollary} \label{corollary2}
With appropriate settings of a learning rate $\eta$ and a constant $C$,
\begin{equation}
    \label{eq:final}
     D_t^{\{T\}}(x) \leq  \sqrt{2T} \eta C \sqrt{\sum_{\tau=t}^{t+T-1}  L_\tau(x)} .
\end{equation}
\end{corollary}

Corollary \ref{corollary2} shows that $\|f(x;w_{t+T}) - f(x;w_{t})\|$ is a lower bound of the square root of accumulated loss $L$ during $T$ gradient descend steps. Thus, when $T$ is fixed, \eg, a certain number of iterations of neural network training, TOD can effectively estimate the loss of sample $x$. Note that the pre-assumptions of Theorem \ref{theorem} and its corollaries limit the learning rate $\eta$ not to be too large to dissatisfy the Taylor expansion used in our proofs. In empirical study we find that the commonly used learning rates, \eg, $\eta$=0.1 or smaller, work well. 
%TOD is also robust to noise since the loss estimation variance is reduced via multiple gradient descend steps. 

\section{Semi-Supervised Active Learning}

\subsection{Problem Formulation}

We first formulate the standard active learning task as follows. Let ($x_S, y_S$) denote a sample pair drawn from a set of labeled data ($X_S, Y_S$), where $X_S$ is the data points and $Y_S$ is the labels. Let $x_U$ denote an unlabeled sample drawn from a larger unlabeled data pool $X_U$, \ie, the labels $Y_U$ corresponding to $X_U$ cannot be observed. In an active learning cycle $c$, the active learning algorithm selects a fixed budget of samples from the unlabeled pool $X_U$ and the selected samples will be annotated by an oracle. The budget size $b$ is usually much smaller than $|X_U|$, the size of the unlabeled pool. The goal of active learning is to select the most informative unlabeled samples for annotation, so as to minimize the expected loss of a task model $f: X \rightarrow Y$. 

%with all labeled data including the newly annotated samples}. 

We next present the use of TOD in a \emph{semi-supervised} active learning framework. An active learning algorithm generally consists of two components: (a) an unlabeled data sampling strategy and (b) the learning of a task model. We adapt TOD to these two components, respectively. For component (a), we propose \name~(\abbr), a new criterion to select unlabeled samples with the largest \emph{estimated loss} for annotation. For component (b), we develop a TOD-based unsupervised loss term to improve the performance of task model. In the following, we formulate the active learning problem and discuss the details of the two components.

%construct $X_B$ by collecting $b$ samples with the largest \abbr~from $X_U$ and obtain their annotations $Y_B$ with an oracle. ($X_B, Y_B$) is then added to labeled pool for the next active learning cycle. As discussed in Theorem \ref{theorem}, the \abbr-based data sampling strategy is able to find samples with the largest loss $L$ in unlabeled pool, so as to minimize the expected loss of model $f$ 

\subsection{\name}

In Eq. \ref{eq:final}, our proposed TOD characterizes a lower bound of the loss function for supervised learning. Here we introduce a variant of TOD, \ie, \name~(\abbr), for active selection of unlabeled samples. \abbr~estimates the sample uncertainty by measuring the difference of model outputs between two consecutive active learning cycles,
\begin{equation}
    D_{\text{cyclic}}(x|w_{c},w_{c-1})=\|f(x;w_{c}) - f(x;w_{c-1})\| ,
\end{equation}
where model parameters $w_{c}$ and $w_{c-1}$ are obtained after the $c$-th and ($c-1$)-th active learning cycle, respectively. 

Fig.~\ref{fig:sampler} illustrates the procedure of COD-based unlabeled data sampling. Given \abbr~for every sample in unlabeled pool $X_U$, our strategy selects $b$ samples with the largest \abbr~from $X_U$. Then, the strategy queries human oracles for annotating the selected samples. The newly-annotated data is added to labeled pool for the next active learning cycle. In the first cycle (\ie, $c=1$), the model $f$ is trained with a random subset of labeled data, and \abbr\ is computed based on the initial model and the model learned after the first cycle. For $c\geq 2$, we compute COD $D_{\text{cyclic}}(x|w_{c},w_{c-1})$ for active sample selection systematically. 
% Since TOD is an effective loss measure, the unlabeled samples with the largest \abbr\ will be selected for oracle annotation. 
\begin{figure}[t]
    \centering
    \includegraphics[width=0.495\linewidth]{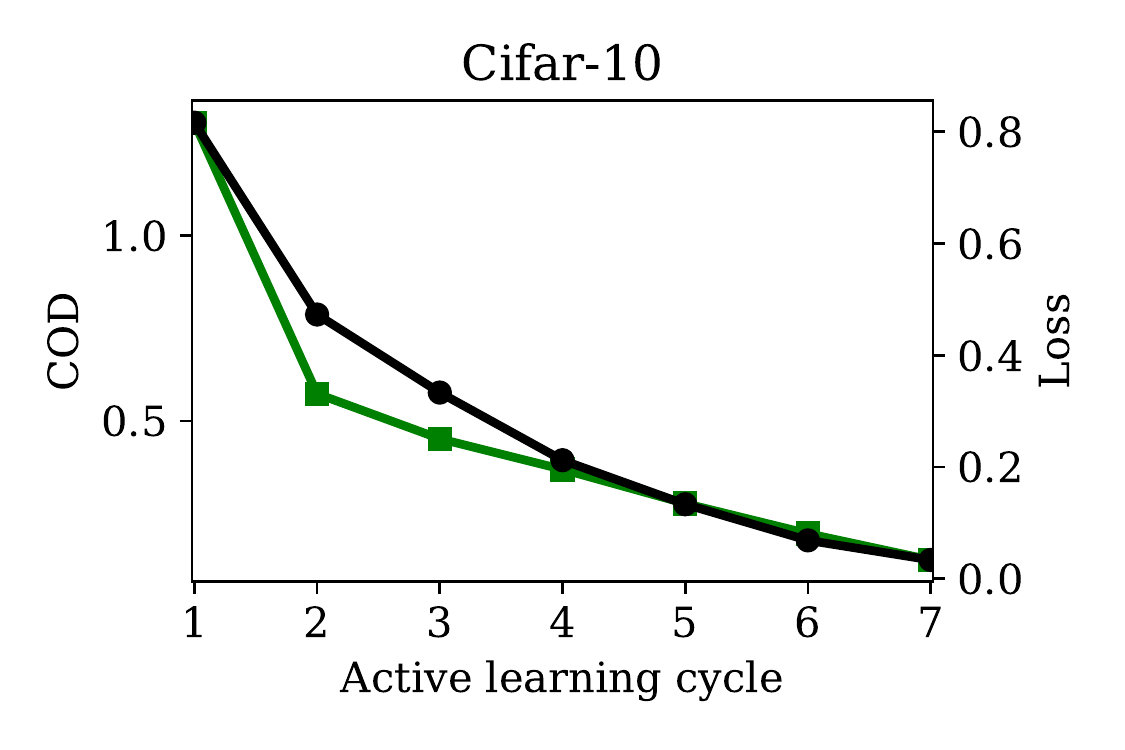}
    \includegraphics[width=0.495\linewidth]{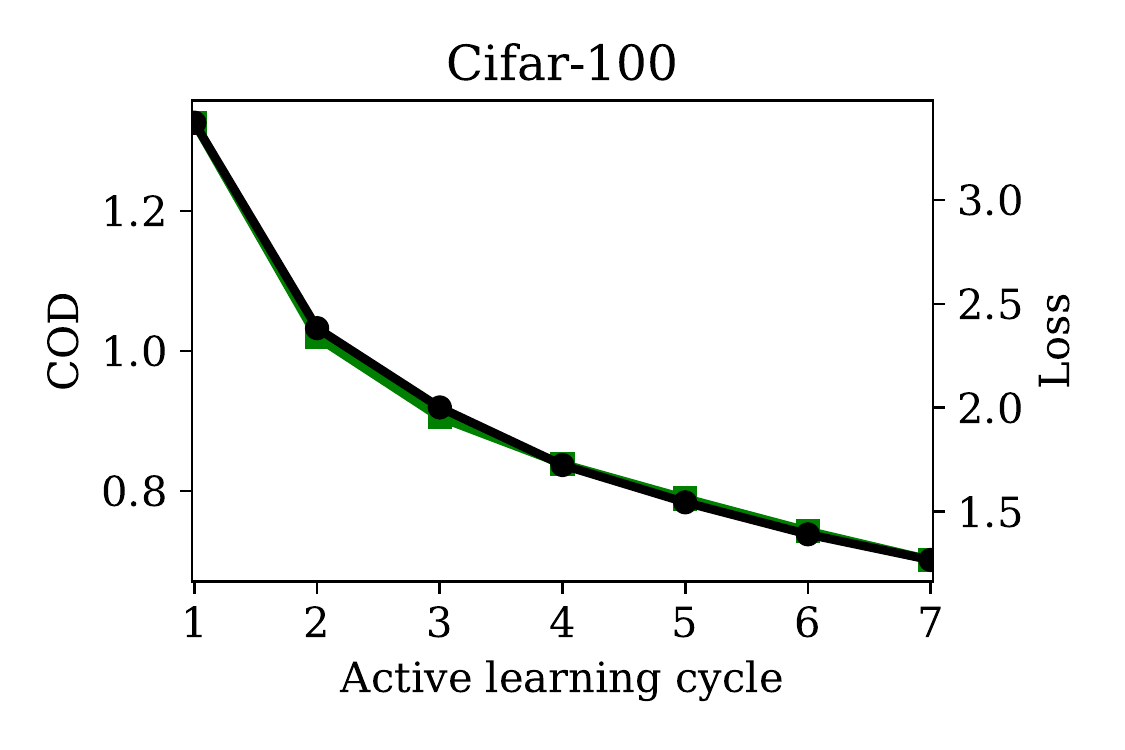}
    \includegraphics[width=0.495\linewidth]{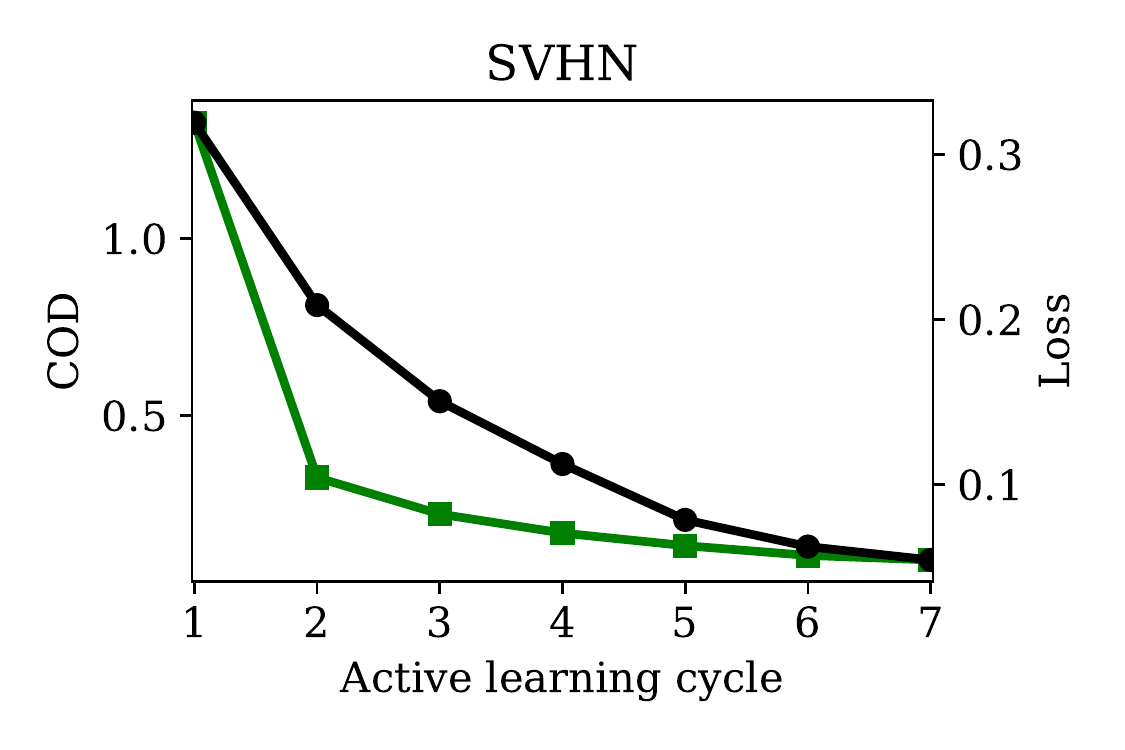}
    \includegraphics[width=0.495\linewidth]{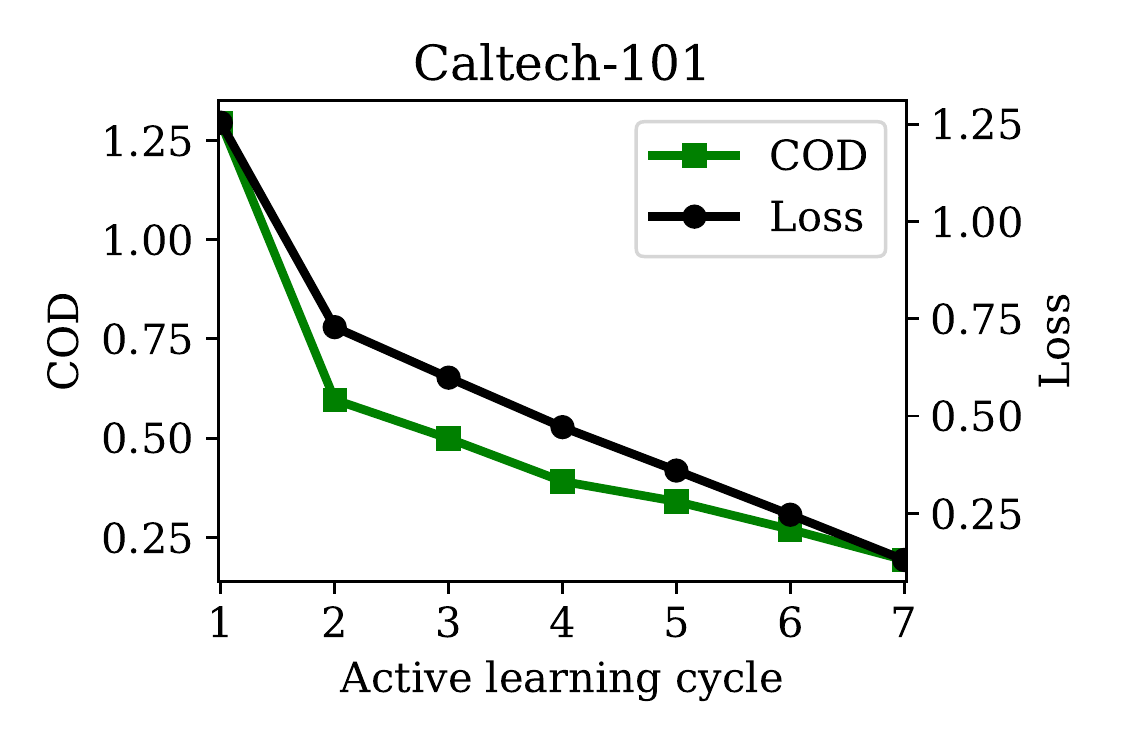}
    \caption{The consistency of \name~(\abbr) and the real task loss. We show COD and real loss averaged over the unlabeled samples, \vs, percentage of labeled images, under active learning setting.}
    \label{fig:distance_vs_cycle}
    \vspace{-.5em}
\end{figure}

\noindent
\textbf{Minimax optimization of COD.} As discussed in Corollary \ref{corollary2}, \abbr-based data sampling strategy can find samples of large loss in unlabeled pool, so as to minimize the expected loss of model $f$ through further training the task model in the next cycle. Fig. \ref{fig:distance_vs_cycle} preliminarily verifies the consistency between \abbr~and the real loss, where \abbr~shows a similar trend with the real loss and they are both decreasing along with the active learning progress. Instead of minimizing the TOD directly (i.e., min-min optimization which may not be good), COD develops TOD as the criterion of sample selection in active learning, where samples with the maximal TOD are picked up (e.g., max-min strategies). When considering labels of samples with potential losses as information gain, our strategy actually maximizes the minimum gain in active learning. 

%\subsection{Output Discrepancy for Task Model Learning}
\subsection{Semi-Supervised Task Learning}
%\abbr~is proposed as an effective data sampling strategy for active learning.
%In addition to the sampling strategy, the proposed algorithm also uses \abbr\ to improve the training procedure of the task model $f$ in a semi-supervised manner. %More specifically, the semi-supervised learning algorithm consists of two components as follows. %(1) an integrated time-evolving loss that combines both supervised and unsupervised tasks based on labeled and unlabeled pools, and (2)
%
%
%\emph{Integrated Time-Evolving Loss.}~

\begin{figure}[t]
    \centering
    \includegraphics[width=0.85\linewidth]{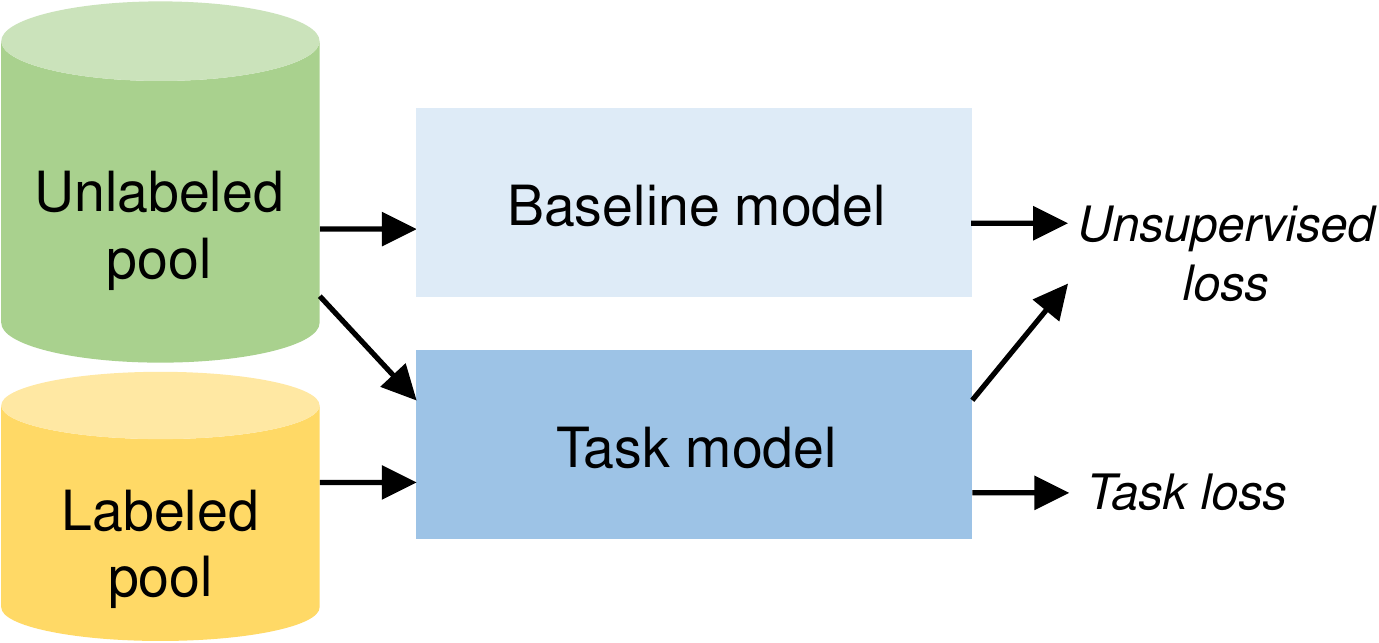}
    \vspace{.5em}
    \caption{Semi-supervised task learning scheme. For labeled data, the task model is trained with the task loss. For unlabeled data, the task model is trained to minimize the distance between outputs of the task model and the baseline model.
    }
    \label{fig:training}
\end{figure}

\noindent
\textbf{Unsupervised loss.} 
As suggested by Corollary \ref{corollary2}, TOD measures the accumulated sample loss, and thus it is natural to employ TOD as an unsupervised criterion to improve the learning of model $f$ using the unlabeled data. However, directly applying TOD to unsupervised training with the baseline model obtained at the last cycle $c-1$ may lead to an unstable training, due to the following aspects: 1) The iteration interval between current model and baseline model (\ie, $T$ in Corollary \ref{corollary2}) is no longer fixed during model training, thus the loss measurement would be inaccurate;
    % The number of SGD iterations in an active learning cycle is often very large in practice, for instance many rounds of iterating over the dataset. 
    %It is Different from the active data sampling in which the current model is always the model learned after a cycle.
2) The baseline model only depends on a single historical model state so that it may suffer from a large variance in loss measurement. To address the above issues, we are inspired by Mean Teacher \cite{emanet} to construct a baseline model by applying an exponential moving average (EMA) to the historical parameters, as
\begin{equation}
    \tilde{w} \gets \alpha\cdot \tilde{w} + (1-\alpha)\cdot w .
\end{equation}
where $\tilde{w}$ and $w$ are parameters of baseline model and current model, respectively, and $\alpha$ is the EMA decay rate. 

Our unsupervised loss minimizes the distance between the current model and the baseline model. In the $c$-th cycle, with the unlabeled pool $X_U^c$, the unsupervised loss is 
\begin{equation}
   L_U^c(w) = \frac{1}{|X^c_U|} \sum_{x_U \in X^c_U} \|f(x_U;w)-f(x_U;\tilde{w})\|^2 .
   %\| f(x_U;w)-f(x_U;\tilde{w})\|^2.
\end{equation}
 %while $w_{t}$ refers to the parameters of the current model in the $t^{th}$ cycle.
 
\noindent
\textbf{Task loss.} 
For the labeled data, we optimize a supervised task objective. Here we take the cross-entropy (CE) loss for image classification as an example. In the $c$-th cycle, given the labeled set $(X^c_S,Y^c_S)$ in the cycle, the supervised loss is
\begin{equation}
    L_S^c(w) = \frac{1}{|X^c_S|}\sum_{(x_S,y_S) \in (X^c_S,Y^c_S)} \mathrm{CE}\left[f(x_S;w),y_S\right] .
\end{equation}
Note that the labeled pool $(X^c_S,Y^c_S)$ will be enlarged per active learning cycle. Within an active learning cycle, the labeled pool remains unchanged. 

\noindent
\textbf{Overall objective.} 
Our semi-supervised task learning scheme is illustrated in Fig. \ref{fig:training}. By integrating the task and unsupervised losses, we minimize an overall learning objective that evolves with the cycle $c$, as 
\begin{equation}
   L_{\text{overall}}^c (w)= L_S^c(w) + \lambda\cdot L_U^c(w),
\end{equation}
where $\lambda$ is a trade-off weight to balance the task and unsupervised loss terms. In our experiments, $\lambda$ is set to 0.05 and the EMA decay rate $\alpha$ is set to 0.999. See supplementary material for more details.

\begin{figure}[t]
    \centering
    \includegraphics[width=0.49\linewidth]{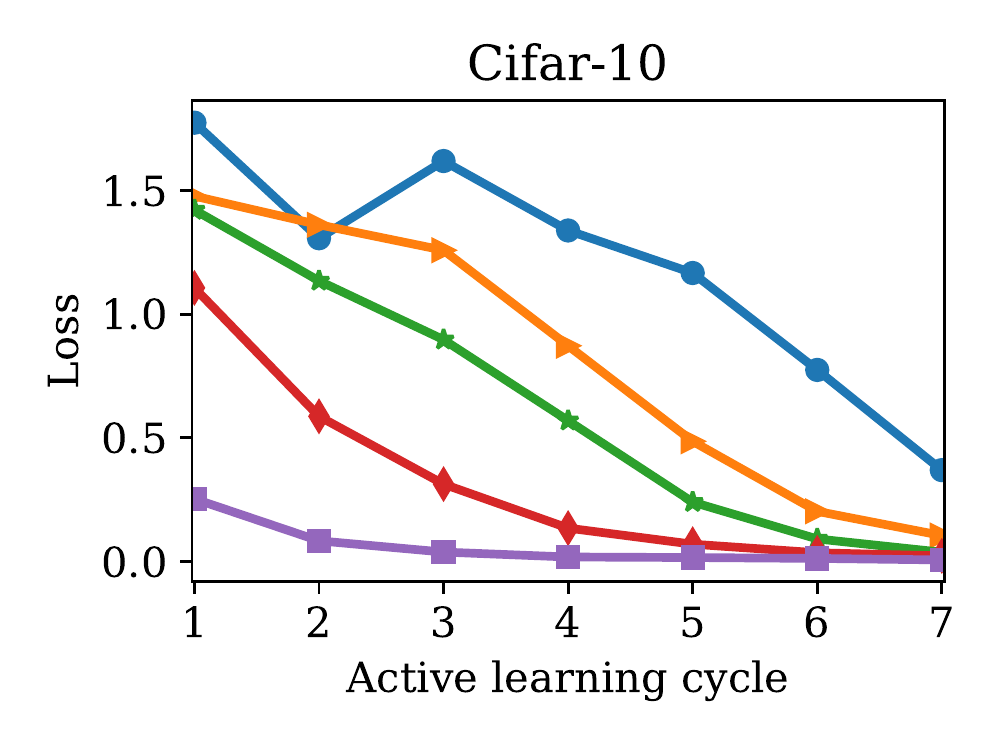}
    \includegraphics[width=0.49\linewidth]{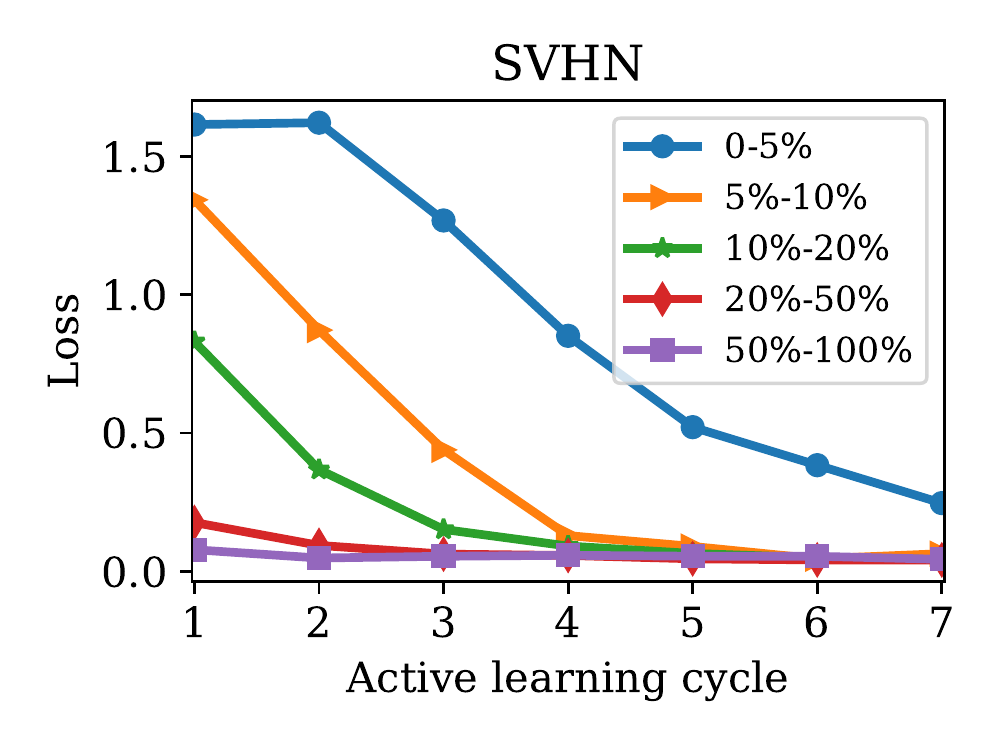}
    \vspace{-.5em}
    \caption{The average real losses of unlabeled samples in a descending order of COD values. For instance, ``0-5\%'' denotes the 5\% unlabeled samples which have the largest COD values, and so on.}
    \label{fig:distance_vs_loss}
    \vspace{-.5em}
\end{figure}

\begin{figure}[t]
    \centering
    \includegraphics[width=0.49\linewidth]{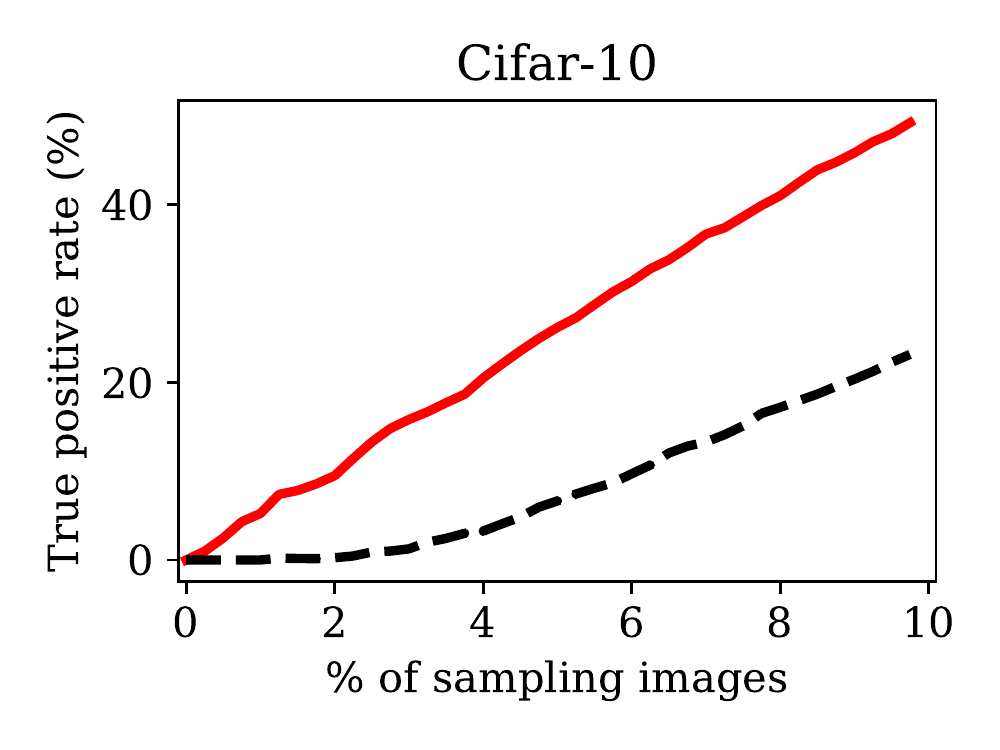}
    \includegraphics[width=0.49\linewidth]{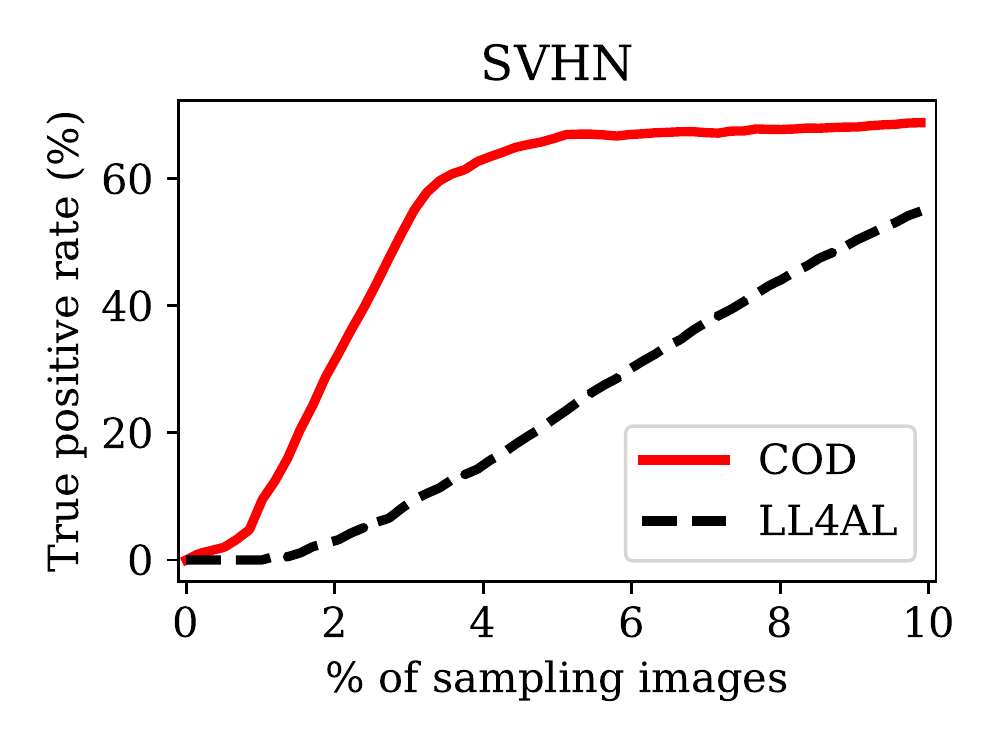}
    \vspace{-.5em}
    \caption{The performance of loss estimation using a learned loss prediction model (LL4AL) \cite{ll4al} and the proposed COD method. We show the proportion of sampled images which have the highest real losses, \vs, the proportion of sampling images. }
    \label{fig:ll4al_vs_rnd}
    \vspace{-.5em}
\end{figure}

\section{Experiments}

We conduct extensive experimental studies to evaluate the proposed active learning approach on two computer vision tasks, image classification and semantic segmentation, with five benchmark datasets. The results are reported over 3 runs with different initial network weights and labeled pools. We implement the methods using PyTorch framework \cite{pytorch}. See supplementary material for more details.

\begin{figure*}[t]
    \vspace{-1em}  
    \centering
    \includegraphics[width=0.36\linewidth]{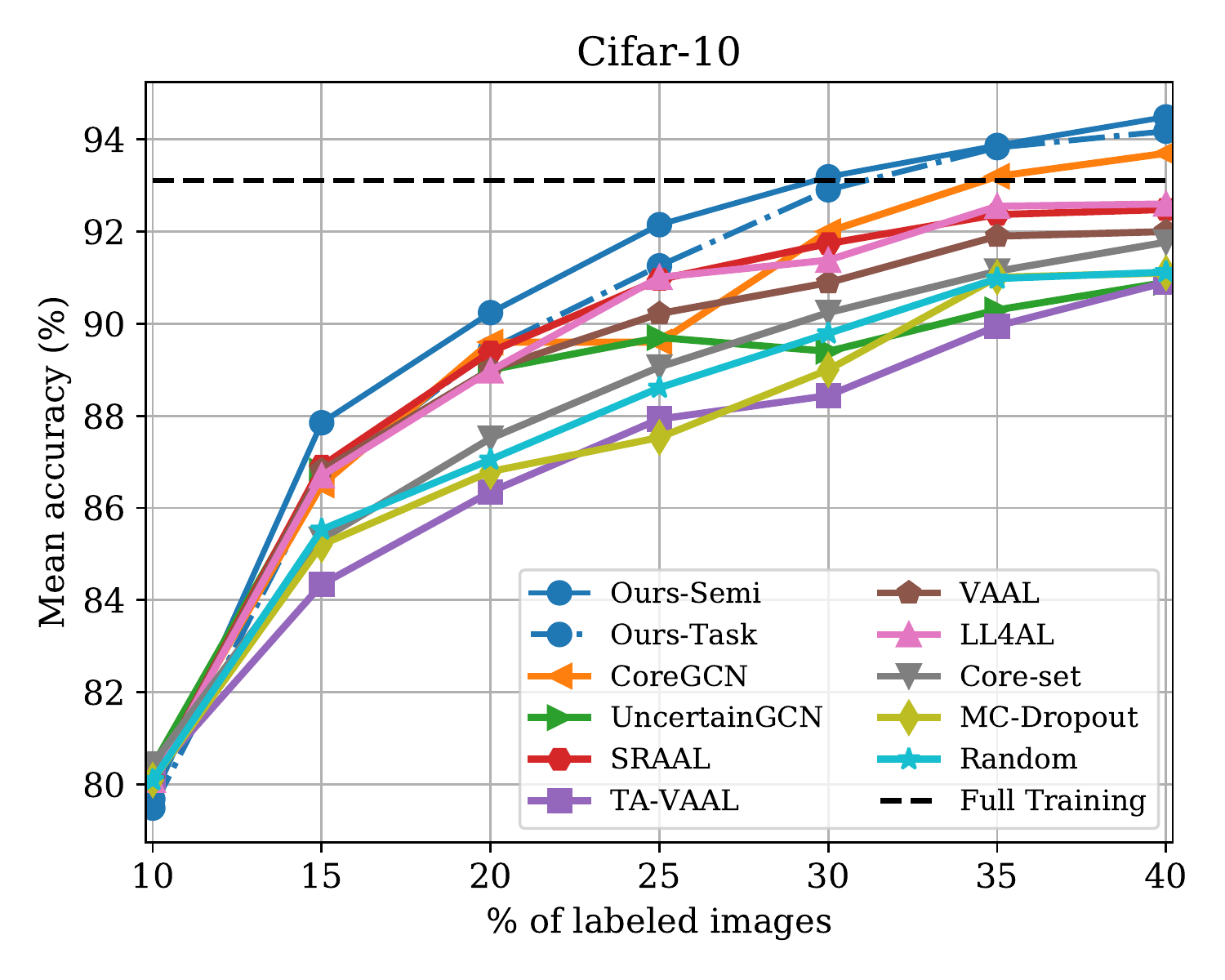}
    \includegraphics[width=0.36\linewidth]{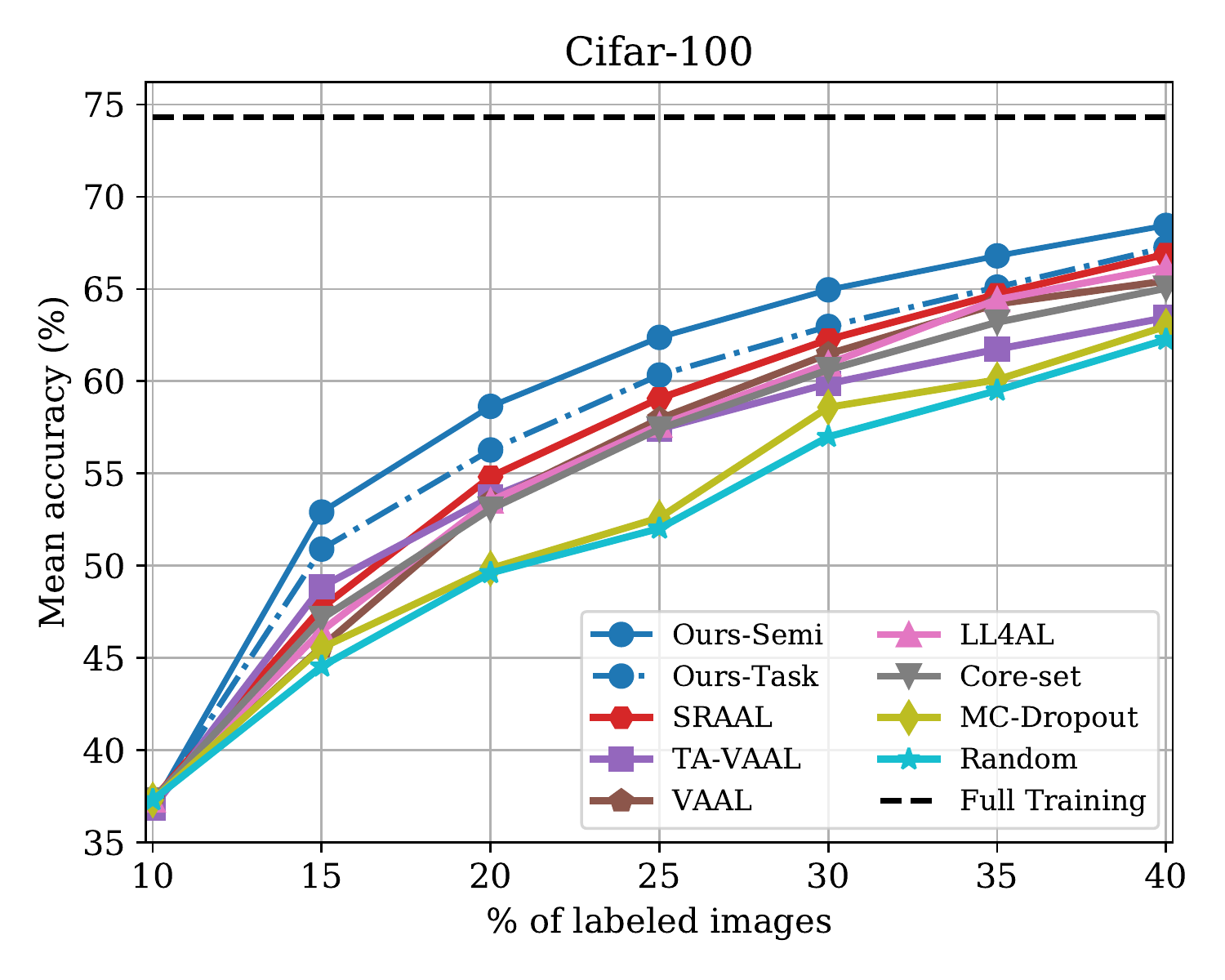}
    \\
    \includegraphics[width=0.36\linewidth]{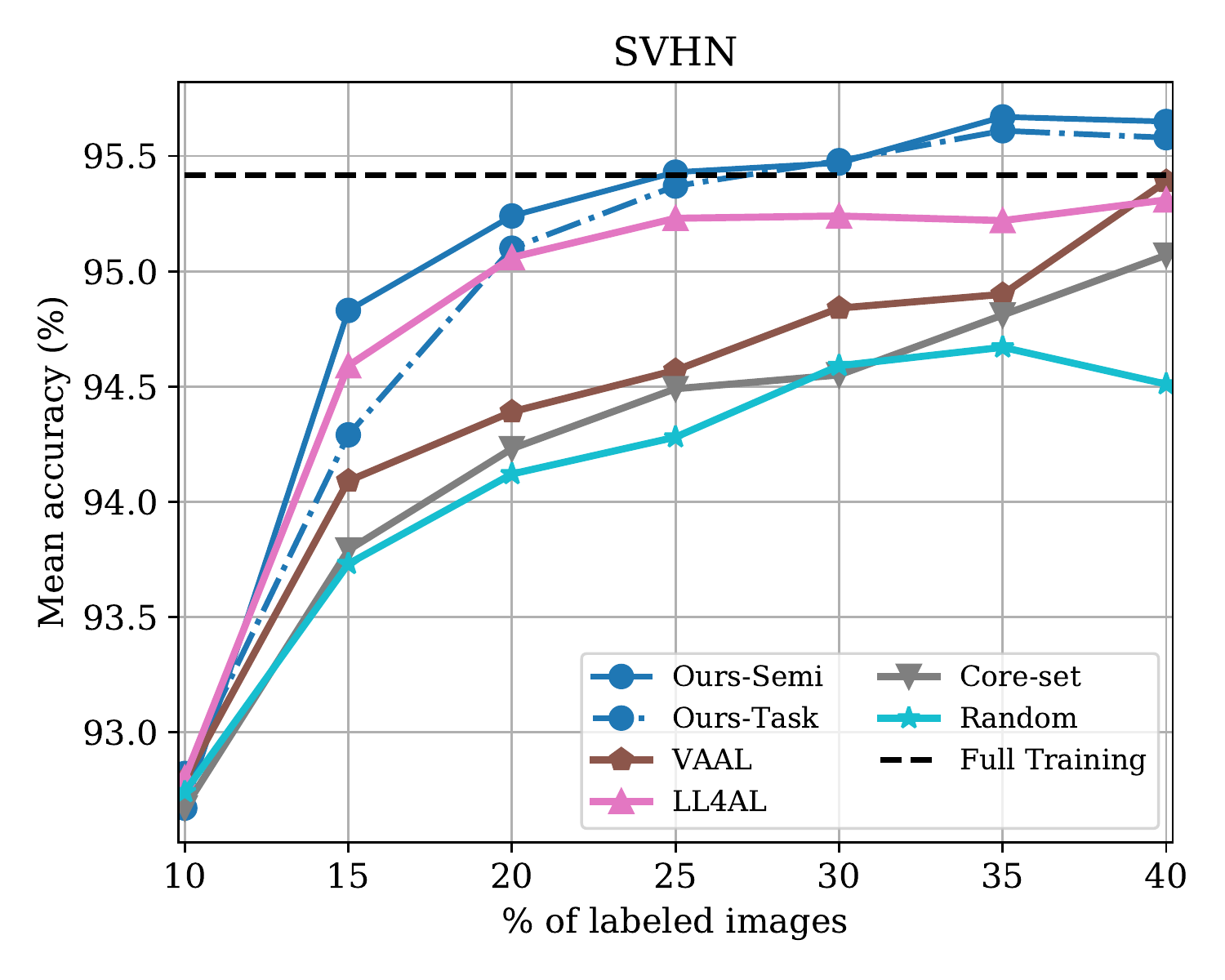}
    \includegraphics[width=0.36\linewidth]{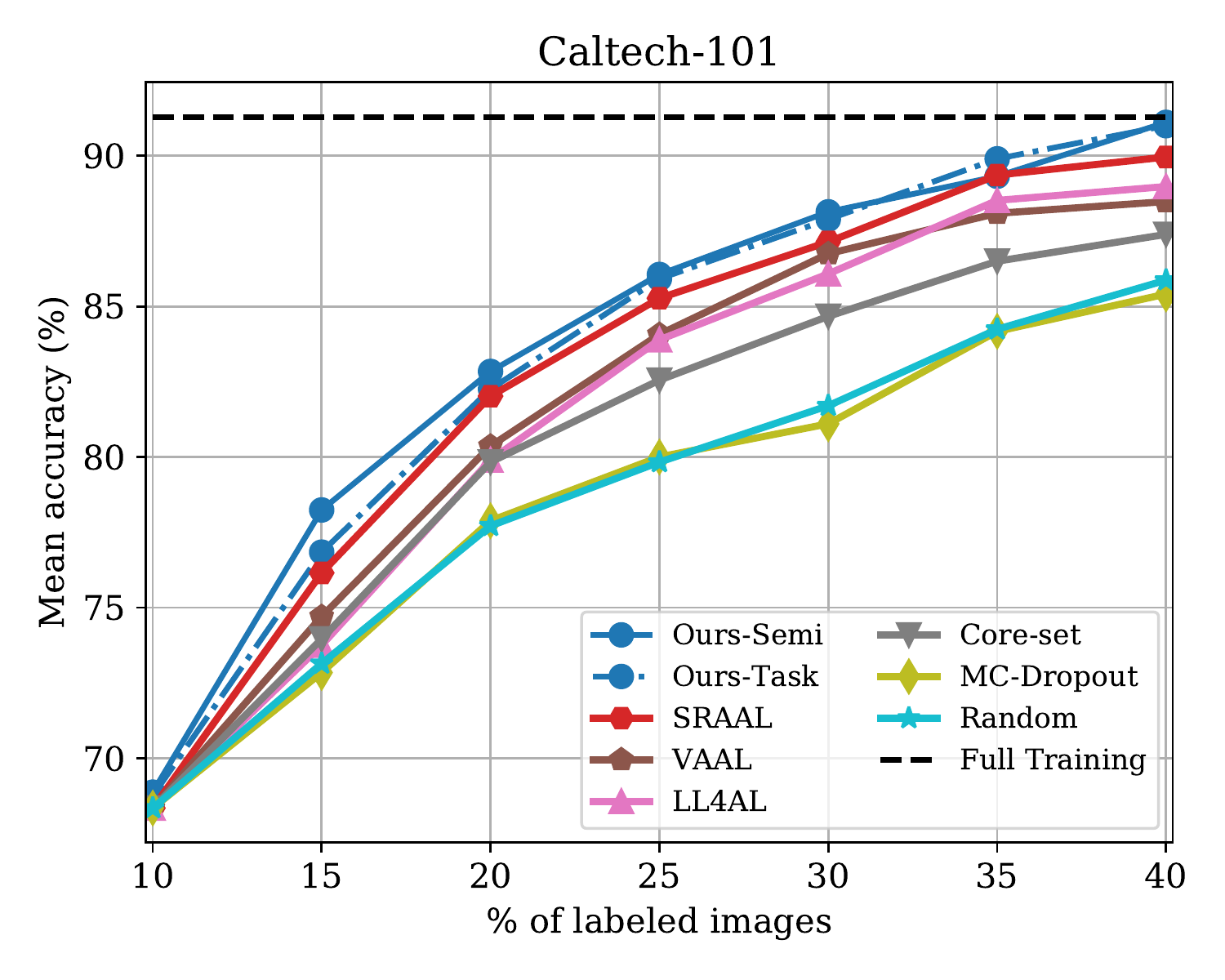}
    \caption{Active learning results of image classification on four benchmark datasets.}
    \vspace{-.5em}
    \label{fig:classification}
\end{figure*}

\subsection{Efficacy of TOD as Loss Measure}
This work proposes TOD to estimate the loss of an unlabeled sample. Fig. \ref{fig:distance_vs_cycle} has evaluated the relations between TOD and sample loss as discussed in Theorem \ref{theorem} and Corollary \ref{corollary2}, suggesting that the average COD and average loss have a consistent trend along with active learning cycles. To further verify the effectiveness of TOD for loss estimation, we study the average loss of unlabeled samples by sorting their COD values. Fig. \ref{fig:distance_vs_loss} shows that the larger COD values of samples indicate the higher losses of samples, and, this observation is consistent across all the active learning cycles. 

In Fig. \ref{fig:ll4al_vs_rnd}, we compare the loss estimation performance of a learned loss prediction model (LL4AL) \cite{ll4al} and COD. We investigate how many samples of the highest losses can be picked out by using different methods. Fig. \ref{fig:ll4al_vs_rnd} shows that COD performs significantly better than LL4AL, as COD is able to pick out more high-loss samples under all the sampling settings. Figs. \ref{fig:distance_vs_cycle}, \ref{fig:distance_vs_loss} and \ref{fig:ll4al_vs_rnd} demonstrate that COD is an effective loss measure as well as a feasible criterion for active data sampling.

\subsection{Active Learning for Image Classification}

\noindent\textbf{Experimental setup.}
We evaluate active learning methods on four benchmark image classification datasets including Cifar-10 \cite{cifar10}, Cifar-100 \cite{cifar10}, SVHN \cite{svhn}, and Caltech-101 \cite{caltech}. Following the conventional practices in deep active learning \cite{ll4al,vaal}, we employ ResNet-18 \cite{resnet} as the image classification model. We compare our active learning approach against the state-of-the-art methods including CoreGCN \cite{coregcn}, UncertainGCN \cite{coregcn}, SRAAL \cite{sraal}, TA-VAAL \cite{tavaal}, VAAL \cite{vaal}, LL4AL \cite{ll4al}, Core-set \cite{coreset}, and MC-Dropout \cite{drop2016}. In addition, the random selection of unlabeled data (``Random'') and the model trained on the full training set (``Full Training'') are also included as baselines. ``Ours-Semi'' indicates our approach trained with the semi-supervised loss and ``Ours-Task'' is our approach trained with only the task loss.

%Due to the limited number of pages, more details of the datasets and our implementations can be found in the appendix.
% We also compare our approach with BDA \cite{bda} and BALD \cite{bald} on SVHN. 
%The results of the compared methods are drawn from the original papers. 

\begin{table*}[t]
\centering
\caption{The active learning performances on 40\% labeled data. `Base': standard task model training without active data selection. `Semi': the proposed semi-supervised task learning. `Active': the proposed active data selection strategy. 
}
\footnotesize
\begin{tabular}{|c||c|c|c|c||c|c|c|c|}
\hline
 Dataset &Core-set & LL4AL & VAAL & SRAAL  & Base & Base+Semi & Base+Active & Base+Semi+Active \\ \hline
Cifar10 & 91.8 & 94.1 & 92.0 & 92.5  & 91.8 & 92.2 (\textcolor{blue}{+0.4}) & \underline{94.2} (\textcolor{blue}{+2.4}) & \textbf{94.5} (\textcolor{blue}{+2.7})    \\\hline
Cifar100 & 65.0 & 65.2 & 65.4 & 66.2 & 62.3 & 66.1 (\textcolor{blue}{+3.8}) & \underline{67.3} (\textcolor{blue}{+5.0}) & \textbf{68.5} (\textcolor{blue}{+6.2})  \\ \hline
\end{tabular}
\label{table:40percent}
\vspace{-.5em}
\end{table*}

\noindent\textbf{Results.}
Fig. \ref{fig:classification} shows image classification performances of different active learning methods. Our method outperforms all the other methods on the benchmark datasets. Additionally, we have the following observations. (i) Our method consistently performs better than the other methods with respect to the cycles. This is a desired property for a successful active learning method, since the labeling budget may vary for different tasks in real-world applications. For instance, one may only be able to annotate 20\% instead of 40\% of all the data. (ii) Our method shows robust performances on difficult datasets such as Cifar-100 and Caltech-101. Both datasets include much more classes than Cifar-10, and, Caltech-101 includes images of much higher resolution (\ie, 300$\times$200). These difficult datasets bring more challenges to active learning, and the superior performances on these datasets demonstrate the robustness of our method. (iii) The performance curves of our method are relatively smooth compared with the other methods. A smooth curve means there are consistent performance improvements from cycle to cycle, indicating that our sampling strategy can take informative data from the unlabeled pool. (iv) Ours-Semi performs better than Ours-Task, demonstrating that our semi-supervised training successfully utilizes the unlabeled data. (v) Our method uses 40\% training samples to outperform the full training on Cifar-10 and SVHN, \eg, 94.5\% \vs 93.1\% on Cifar-10. This interesting finding is in accord with the observations discussed in previous literature \cite{koh2017understanding} that some data in the original dataset might be unnecessary or harmful to model training. 
%In later cycles, Ours-Semi shows a relatively larger accuracy improvement. We conjecture this

Table \ref{table:40percent} compares different active learning methods, \ie, the state-of-the-art algorithms and the proposed methods, for image classification on 40\% training labeled data. Both semi-supervised task learning and active data selection strategy contributes to performance improvement, while, active data selection results in a more significant improvement than semi-supervised task learning. We also note that the proposed method can outperform existing algorithms without semi-supervised task learning (see `Base+Active' in Table \ref{table:40percent}).

%It indicates that our data sampling strategy based on the cyclic discrepancy is effective.
% Specifically, the proposed strategy performs data sampling in a greedy fashion. The data sampled in the current cycle only depends on the current trained model and the model trained in the immediate previous cycle. 
% It could be feasible to train a deep learning model with a subset of the original dataset, without sacrificing the performance. 
% {\sout{In addition to effective data sampling, involving unlabeled data in training makes more contributions to the performance improvement in the later cycles.}} 
% {\sout{Our method could be feasible to train a model on a subset of the original dataset without sacrificing the performance.}} 

\begin{figure}[t]
    \vspace{-.5em}  
    \centering
    \includegraphics[width=0.85\linewidth]{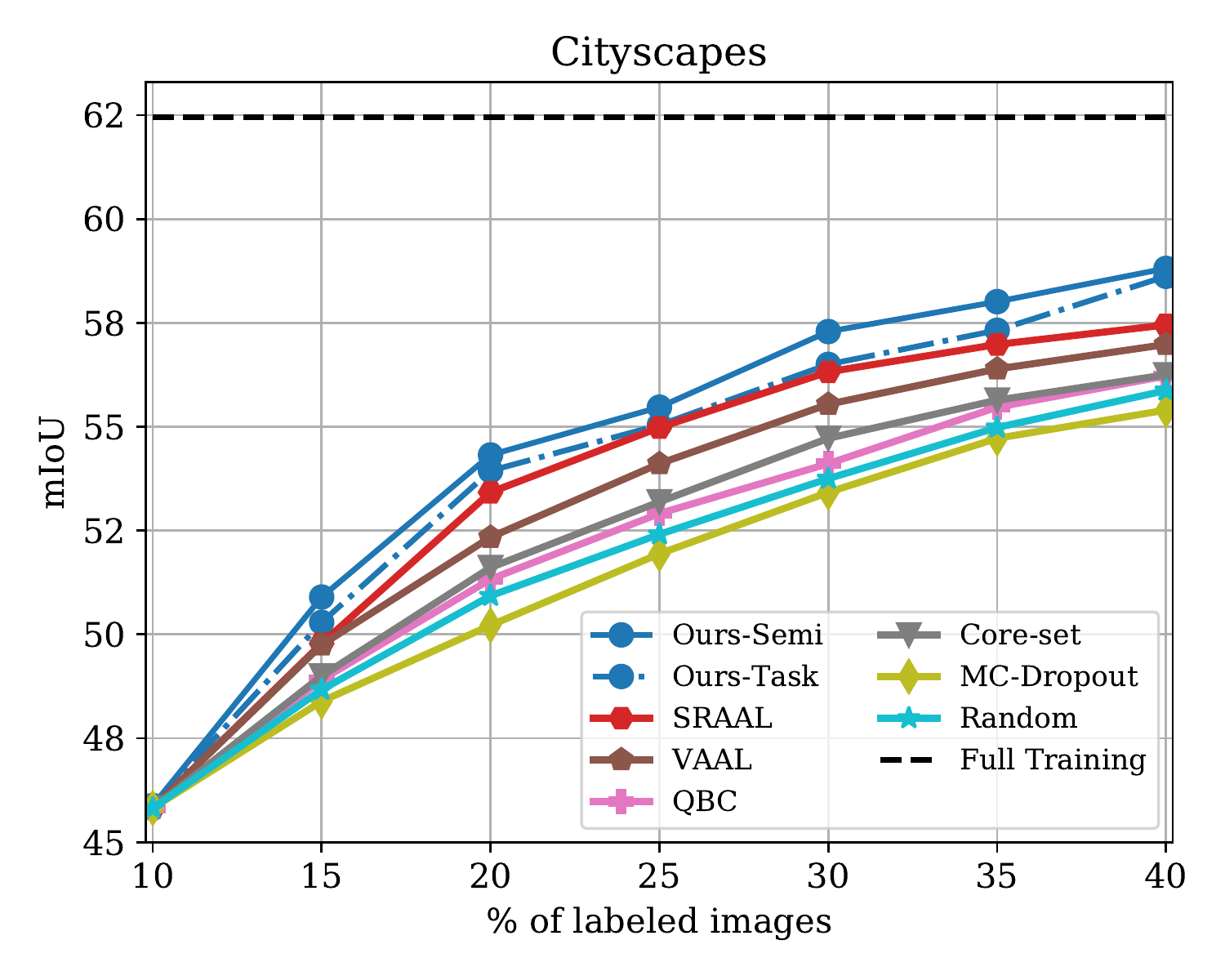}
    \caption{Active learning results of semantic segmentation on CityScapes dataset.
    }
    \vspace{-.5em}
    \label{fig:segmentation}
    \vspace{-.5em}
\end{figure}

\subsection{Active Learning for Semantic Segmentation}

\noindent\textbf{Experimental setup.}
To validate the active learning performance on more complex and large-scale scenarios, we study the semantic segmentation task with the Cityscapes dataset \cite{cityscapes} which is a large-scale driving video dataset collected from urban street scenes. Semantic segmentation addresses the pixel-level classification task and its annotation cost is much higher. Following the settings in \cite{vaal,sraal}, we employ the 22-layer dilated residual network (DRN-D-22) \cite{drn} as the semantic segmentation model. We report the mean Intersection over Union (mIoU) on the validation set of Cityscapes. We compare our method against SRAAL \cite{sraal}, VAAL \cite{vaal}, QBC \cite{hybrid2018}, Core-set \cite{coreset}, MC-Dropout \cite{drop2016}, and the random selection. 

\noindent\textbf{Results.}
Fig. \ref{fig:segmentation} shows the semantic segmentation performances of different active learning methods on Cityscapes. Both Ours-Semi and Ours-Task outperform the other baselines in terms of mIoU. The results demonstrate the competence of our approach on the challenging semantic segmentation task. Note that in our approach, neither task model training nor data sampling needs to exploit extra domain knowledge. Therefore, our approach is independent of tasks. Moreover, the image size of Cityscapes (\ie, 2048$\times$1024) is much larger than that of the classification benchmarks, indicating that our method is not sensitive to the data complexity. These advantages make our approach a competitive candidate for complex real-world applications.

% our output discrepancy-based semi-supervised learning
%On the other hand, this experiment indicates the 'global view' of our method, and it is not biased towards some classes only. It is worth noting that the term 'global view' does not contradict with the 'greedy search' mentioned in the Results section of Image Classification. The 'global view' refers to global data, whereas 'greedy search' refers to the data sampling fashion that is only based on current trained model and the model trained in its immediate previous cycle.  

\begin{figure}[t]
    \centering
    \includegraphics[width=0.49\linewidth]{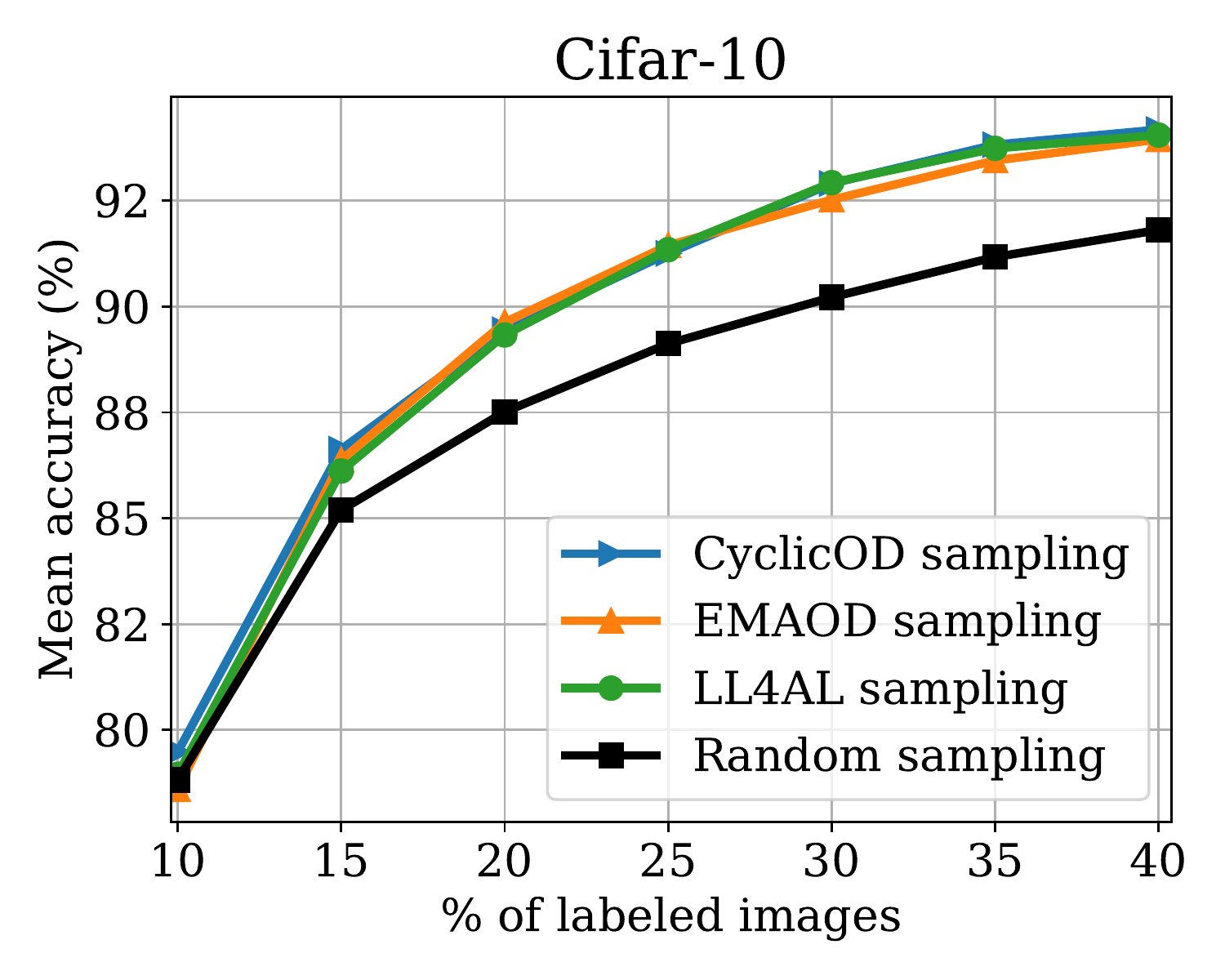}
    \includegraphics[width=0.49\linewidth]{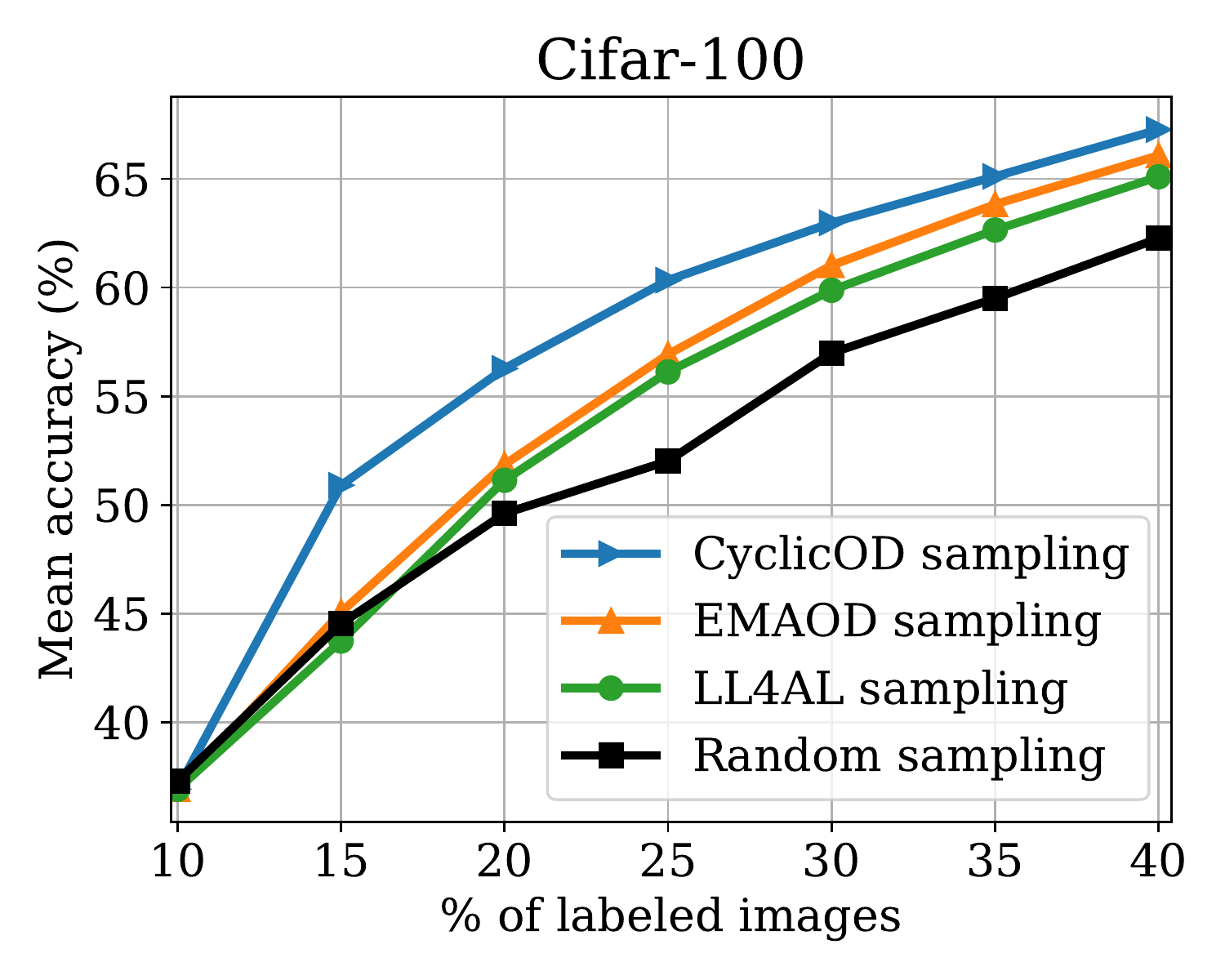}
    \vspace{-.2em}
    \caption{Ablation on active data sampling. %We compare the full active learning pipeline, two variants of the output discrepancy-based sampling (CyclicOD and EMAOD), LL4AL sampling \cite{ll4al}, and random sampling.
    }
    \label{fig:sampling}
    \vspace{-.5em}
\end{figure}

\begin{figure}[t]
    \centering
    \includegraphics[width=0.49\linewidth]{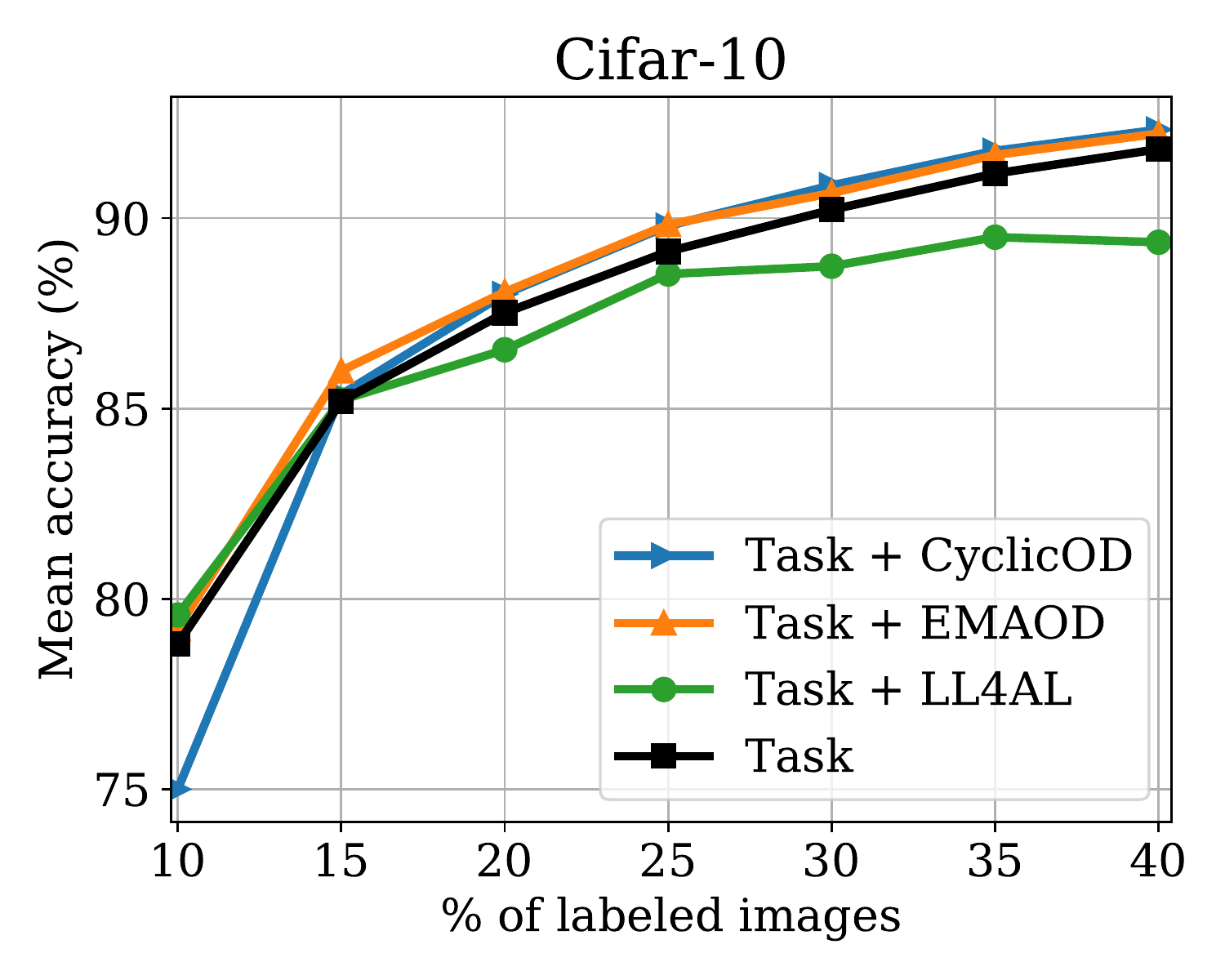}
    \includegraphics[width=0.49\linewidth]{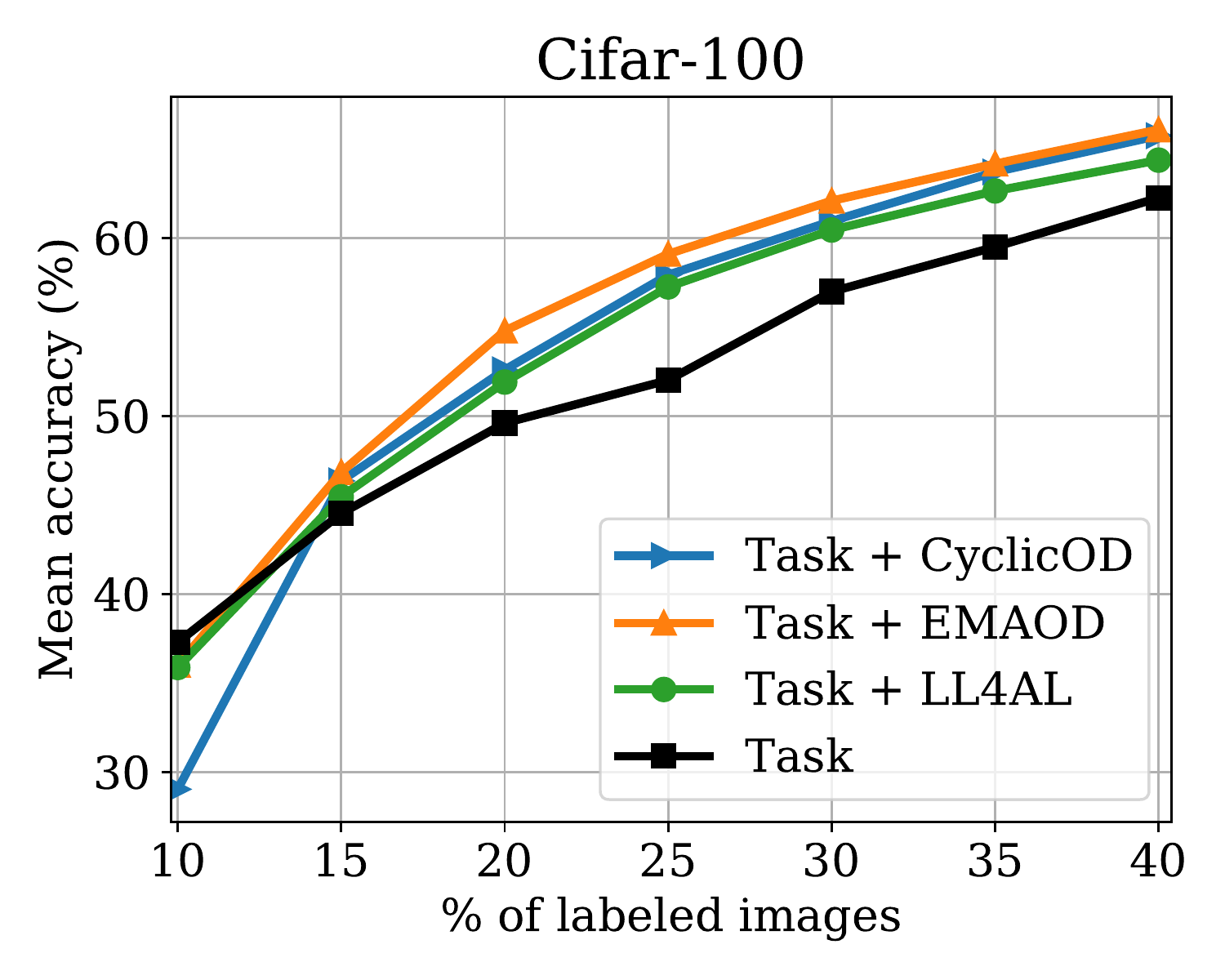}
    \vspace{-.2em}
    \caption{Ablation on semi-supervised task learning. 
    }
    \label{fig:unsupervised}
    \vspace{-1em}
\end{figure}

\begin{table*}[t]
\centering
\caption{The class-wise performances on Cityscapes, where 40\% labeled data is used for training. `Proportion' denotes the proportions of classes at pixel level. `T' is the model trained using only the task loss. `T + U' is the model trained using both the task loss and the proposed TOD-based unsupervised loss. 
}
\scriptsize
\begin{tabular}{|l|ccccccccccccccccccc|c|}
\hline
\textbf{Class ID} & 1 & 2 & 3 & 4 & 5 & 6 & 7 & 8 & 9 & 10 & 11 & 12 & 13 & 14 & 15 & 16 & 17 & 18 & 19 & \textbf{Ave} \\
\hline
\textbf{Proportion} (\%) & 37.4 & 5.4 & 22.3 & 0.8 & 0.8 & 1.5 & 0.2 & 0.7 & 17.3 & 0.8 & 3.3 & 1.3 & 0.2 & 6.5 & 0.3 & 0.4 & 0.1 & 0.1 & 0.7 & - \\

\textbf{T} (mIoU) & 92 & 67 & 82 & 16 & 27 & 53 & 53 & 63 & 87 & 42 & \textbf{84} & 71 & 43 & 86 & 19 & 34 & \textbf{20} & 31 & 69 & 54.7 \\

\textbf{T + U} (mIoU) & \textbf{95} & \textbf{72} & \textbf{87} & \textbf{24} & \textbf{28} & \textbf{56} & \textbf{59} & \textbf{72} & \textbf{90} & \textbf{49} & \textbf{84} & \textbf{77} & \textbf{52} & \textbf{90} & \textbf{24} & \textbf{42} & 8 & \textbf{39} & \textbf{73} & \textbf{58.9} \\
\hline
\end{tabular}
\label{table:class}
\end{table*}

\subsection{Ablation Study}
%We conduct more ablation studies to investigate the role of each module in the proposed method. We validate the competence of our data sampling strategy and the necessity of exploiting unsupervised learning in active learning. We also perform empirical studies on the hyper-parameter selection.

% Thirdly, we suggest optimal hyper-parameters for the use of EMA, based on our experimental observations. 
% in order to enhance the task model training (\eg, classification)

\noindent
\textbf{Active data sampling strategy.}
Fig. \ref{fig:sampling} compares different active data sampling strategies on Cifar-10 and Cifar-100. CyclicOD and EMAOD are two variants of TOD, where CyclicOD employs the model at the end of last cycle as the baseline model while EMAOD employs an exponential moving average of the previous models as the baseline model. LL4AL \cite{ll4al} uses a learned loss prediction module to sample the unlabeled data.  Fig. \ref{fig:sampling} shows that the proposed sampling strategies, \ie, EMAOD and CyclicOD, outperform random sampling and LL4AL sampling on both datasets, validating the effectiveness of TOD-based sampling strategy. CyclicOD performs better than EMAOD on Cifar-100, thus we employ COD as our sampling strategy in the rest of the experiments.
% The full pipeline is our proposed active learning approach, consisting of the CyclicOD-based data sampling and the semi-supervised model learning.
%, and our full pipeline achieves the best performance
%  as well as the semi-supervised learning framework

\noindent
\textbf{Semi-supervised task learning.}
To evaluate the necessity of semi-supervised task learning in active learning, Fig. \ref{fig:unsupervised} compares different loss functions on Cifar-10 and Cifar-100. CyclicOD loss and EMAOD loss are two TOD-based unsupervised learning criteria. They are minimized on the unlabeled data, and, the settings of their baseline models are identical to those in the study of sampling strategy as discussed above. LL4AL loss \cite{ll4al} minimizes the distance between the predicted loss and the real task loss, and it needs the data labels. All the auxiliary losses are used in a combination with the task loss. The full pipeline and the training with only the task loss are also included in comparison. We observe that either the EMAOD loss or the CyclicOD loss can help to improve the performance, and either of them shows a larger performance improvement than the LL4AL loss. The EMAOD loss demonstrates a more stable performance than the CyclicOD loss, indicating that directly applying COD~to unsupervised training may lead to an unstable model training. A moving average of previous model states enables a more stable unsupervised training. We employ EMAOD as our unsupervised loss in the rest of the experiments.

Table \ref{table:class} shows the per-class performance of standard task model training on Cityscapes, where 40\% labeled data is observable. The row of `Proportion' in Table \ref{table:class} shows the pixel-level proportion of every class, indicating a severe class imbalance problem of Cityscapes. We compare the models trained without (\ie, `T') and with (\ie, `T + U') the unsupervised loss. The semi-supervised learning yields better results on 18 out of 19 classes. More importantly, the semi-supervised learning shows more significant performance improvements on the minority classes than the majority classes, demonstrating that the unsupervised loss imparts robustness to the task model to handle the class imbalance issue.

\begin{figure}[t]
    \vspace{-.5em}  
    \centering
    \includegraphics[width=0.7\linewidth]{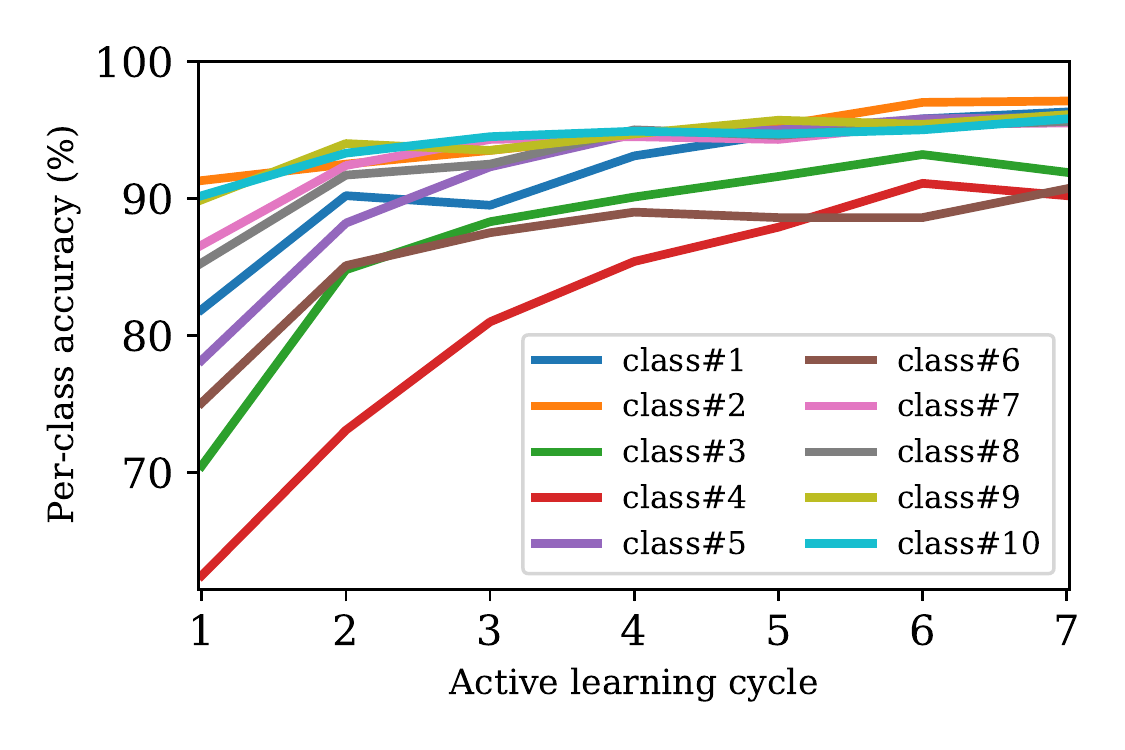}
    \vspace{-.5em}
    \caption{Per-class accuracy on Cifar-10 using the proposed active learning method.}
    \label{fig:per_class}
\end{figure}

\noindent
\textbf{Per-class performance.}
Fig. \ref{fig:per_class} shows the per-class accuracy on Cifar-10 with the proposed active learning method. The accuracies of the classes are improved along with the increasing of active learning cycles in most cases, such that the performance improvement is not biased towards certain classes. The accuracies of class\#3 and class\#4 decrease from the 6-th cycle to the 7-th cycle, mainly due to the overfitting. 

\begin{table}[t]
\centering
\caption{Time (seconds) taken for one iteration of active sampling using an NVIDIA GTX 1080Ti GPU. \textit{Number of sampling images} and \textit{size of images} are shown for each dataset.
}
\footnotesize
\begin{tabular}{|c|c|c|c|c|}
\hline
\multirow{2}{*}{Method} & \multirow{2}{*}{\begin{tabular}[c]{@{}c@{}}Cifar-10\\ \it 2.5K, $32^2$ \end{tabular}} & \multirow{2}{*}{\begin{tabular}[c]{@{}c@{}}SVHN\\ \it 3.6K, $32^2$ \end{tabular}} & \multirow{2}{*}{\begin{tabular}[c]{@{}c@{}}Caltech-101\\ \it 0.4K, $224^2$ \end{tabular}} & \multirow{2}{*}{\begin{tabular}[c]{@{}c@{}}Extra\\ model?\end{tabular}} \\ & & & &  \\ \hline
Coreset \cite{coreset}     & 91.4 & 168.7       & 48.2 & $\times$      \\
VAAL  \cite{vaal}& 13.0      & 17.2  & 32.6  &  $\surd$ \\ 
LL4AL \cite{ll4al}& 7.7       & 10.8  & 39.6  &  $\surd$        \\ \hline
COD (ours)  & \textbf{7.2}       & \textbf{10.1}  & \textbf{26.9}  & $\times$        \\ \hline
\end{tabular}
\label{table:time}
\end{table}

\subsection{Time Efficiency}

This paper proposes to use COD for active data sampling. Table \ref{table:time} evaluates the time taken for one active sampling iteration using different active learning methods. On all the three image classification datasets with different number and size of sampling images, COD is faster than the existing active learning methods. COD is task-agnostic and more efficient, since it only relies on the task model itself and it does not introduce extra learnable models such as the adversarial network (VAAL) \cite{vaal} or the loss prediction module (LL4AL) \cite{ll4al}.

\section{Conclusion}
In this paper we have presented a simple yet effective deep active learning approach. The core of our approach is a measurement Temporal Output Discrepancy~(TOD) which estimates the loss of unlabeled samples by evaluating the discrepancy of outputs given by models at different gradient descend steps. We have theoretically shown that TOD~lower-bounds the accumulated sample loss. On basis of TOD, we have developed an unlabeled data sampling strategy and a semi-supervised training scheme for active learning. Due to the simplicity of TOD, our active learning approach is efficient, flexible, and easy to implement. Extensive experiments have demonstrated the effectiveness of our approach on image classification and semantic segmentation tasks. In future work, we plan to apply TOD to other machine learning tasks and scenarios, as it is an effective loss measure.

{\small
\bibliographystyle{ieee_fullname}
\bibliography{main}
}

\clearpage
\appendix

\setcounter{theorem}{0}
\setcounter{remark}{0}
\setcounter{corollary}{0}
% \setcounter{equation}{0}
% \makeatletter 
% \renewcommand{\thefigure}{A\@arabic\c@figure}
% \renewcommand{\thetable}{A\@arabic\c@table}
% \makeatother

\begin{center}
\Large{\textbf{Supplementary Material}}
\end{center}

\section{Proofs}
\begin{theorem} 
With an appropriate setting of learning rate $\eta$,
\begin{equation}
    \label{eq:theorem1}
     D_t^{\{1\}}(x) \leq \eta \sqrt{2L_t(x)} \| \nabla_{w} f(x;w_t)\|^2 . 
\end{equation}
\end{theorem}

\noindent\textit{Proof.} We apply one-step gradient descent to $w_t$ then using first-order Taylor series,
\begin{align}
    \label{eq:one_step}
    D_t^{\{1\}}(x) \overset{\text{def}}{=} ~& \|f(x;w_{t+1}) - f(x;w_{t})\|   \\ \notag
    = ~& \|f(x;w_{t}-\eta\nabla_{w_t} L_t(x)) - f(x;w_{t})\| \\ \notag
    = ~& \|f(x;w_{t}) - \eta\nabla_wf(x;w_t)^\mathrm{T} \nabla_w L_t(x) - f(x;w_{t})\|\\ \notag
    = ~& \|- \eta\nabla_wf(x;w_t)^\mathrm{T}\nabla_w L_t(x)\|.
\end{align}
Recall that 
\begin{align}
    \label{eq:nablalt}
    \nabla_w L_t(x) = (y-f(x;w_t))\cdot \nabla_w f(x;w_t).
\end{align}
By substituting Eq. \ref{eq:nablalt} into Eq. \ref{eq:one_step},
\begin{align}
    D_t^{\{1\}}(x) = ~& \eta \|(y-f(x;w_t)) \cdot \nabla_w f(x;w_t)^\mathrm{T} \nabla_w f(x;w_t) \| \\ \notag
    \leq ~& \eta \|(y-f(x;w_t))\| \cdot  \| \nabla_{w} f(x;w_t)\|^2  \\ \notag
    = ~& \eta \sqrt{2L_t(x)} \| \nabla_{w} f(x;w_t)\|^2 . \\ \notag
\end{align}

\begin{corollary} 
With an appropriate setting of learning rate $\eta$,
\begin{equation}
    \label{eq:T_step}
     D_t^{\{T\}}(x) \leq  \sqrt{2} \eta \sum_{\tau=t}^{t+T-1} \left( \sqrt{L_\tau(x)} \| \nabla_{w} f(x;w_\tau)\|^2 \right) . 
\end{equation}
\end{corollary}

\noindent\textit{Proof.} 
\begin{align}
    \label{eq:T_step1}
     D_t^{\{T\}}(x) \overset{\text{def}}{=} ~&\|f(x;w_{t+T}) - f(x;w_{t})\| \\ \notag
     \leq ~& \sum_{\tau=t}^{t+T-1} \|  f(x;w_{\tau+1}) - f(x;w_{\tau}) \| \\ \notag 
     \leq ~& \sqrt{2} \eta \sum_{\tau=t}^{t+T-1} \left( \sqrt{L_\tau(x)} \| \nabla_{w} f(x;w_\tau)\|^2 \right) . \\ \notag
\end{align}

\begin{remark}
For a linear layer $\phi(x;W)$ with ReLU activation, the Lipschitz constant $\mathcal{L}(W) \leq \|x\|$.
\end{remark}
\noindent\textit{Proof.} 
\begin{align}
   & \|\phi(x;W+r)-\phi(x;W)\| \\ \notag
   = ~& \| \max(0, (W+r)^\mathrm{T}x+b) - \max(0, W^\mathrm{T}x+b)\| \\ \notag
   \leq ~& \|r^\mathrm{T}x\| \\ \notag 
   \leq ~& \|x\| \cdot \|r\| .
\end{align}
Therefore, the Lipschitz constant $\mathcal{L}(W) \leq \|x\|$.

\begin{corollary} 
With appropriate settings of a learning rate $\eta$ and a constant $C$,
\begin{equation}
    \label{eq:final}
     D_t^{\{T\}}(x) \leq  \sqrt{2T} \eta C \sqrt{\sum_{\tau=t}^{t+T-1}  L_\tau(x)} .
\end{equation}
\end{corollary}
\noindent\textit{Proof.} 
By substituting $\| \nabla_w f \|^2 \leq C$ into Corollary \ref{corollary1} then applying Cauchy–Schwarz inequality, we have
\begin{align}
    \label{eq:Cauchy–Schwarz}
     D_t^{\{T\}}(x) 
     \leq ~& \sqrt{2} \eta C \sum_{\tau=t}^{t+T-1} \sqrt{L_\tau(x)}  \\ \notag 
     \leq ~& \sqrt{2T} \eta C \sqrt{\sum_{\tau=t}^{t+T-1}  L_\tau(x)} .
\end{align}

\begin{table*}[!htb]
\centering
\caption{The summary of datasets used in the experiments. `\#classes' and `image size' are characters after pre-processing. `image size' is the size of images used for training. 
}
\begin{tabular}{|l|c|c|c|c|c|c|c|c|c|}
\hline
dataset & task & content & \#classes & image size & train & val & test \\
\hline\hline
Cifar-10  & image classification & natural images & 10 & 32$\times$32 & 45,000 & 5,000 & 10,000 \\ \hline
Cifar-100 & image classification & natural images & 100 & 32$\times$32 & 45,000 & 5,000 & 10,000 \\ \hline
SVHN  & image classification & street view house numbers & 10 & 32$\times$32 & 65,931 & 7,326 & 26,032 \\ \hline
Caltech-101 & image classification & natural images & 101 & 224$\times$224 & 7,316 & 915 & 915 \\ \hline
Cityscapes  & semantic segmentation & driving video frames & 19 & 688$\times$688 & 2,675 & 300 & 500 \\ \hline
\end{tabular}
\label{appendix:dataset}
\end{table*}

\begin{table*}[!htb]
\centering
\caption{The summary of implementation details on each dataset. `start' is the number of initially labeled samples and `budget' is the number of newly annotated samples in each cycle. `cycle' is the number of active learning cycles. $\alpha$ is the EMA decay rate and $\lambda$ is the weight for unsupervised loss.
}
\begin{tabular}{|l|c|c|c|c|c|c|c|c|c|c|c|}
\hline
dataset & start & budget & cycle & optimizer & lr & momentum & decay & epochs  & batch & $\alpha$ & $\lambda$   \\
\hline\hline
Cifar-10  & 10\% & 5\% & 7 & SGD & 0.1 & 0.9 & 5$\times$10$^{-4}$ & 200  & 128 & 0.999 & 0.05 \\ \hline
Cifar-100 & 10\% & 5\% & 7 & SGD & 0.1 & 0.9 & 5$\times$10$^{-4}$ & 200  & 128 & 0.999 & 0.05 \\ \hline
SVHN  & 10\% & 5\% & 7 & SGD & 0.1 & 0.9 & 5$\times$10$^{-4}$ & 200  & 128 & 0.999 & 0.05 \\ \hline
Caltech-101 & 10\% & 5\% & 7 & SGD & 0.01 & 0.9 & 5$\times$10$^{-4}$ & 50  & 64 & 0.999 & 0.05 \\ \hline
Cityscapes & 10\% & 5\% & 7 & Adam & 5$\times$10$^{-4}$ & - & - & 40 & 4 & 0.999 & 0.05 \\ \hline
\end{tabular}
\label{appendix:learning}
\end{table*}

\section{Experimental Details}

\subsection{Image Classification}

\paragraph{Datasets}
We evaluate the active learning methods on four common image classification datasets, including Cifar-10 \cite{cifar10}, Cifar-100 \cite{cifar10}, SVHN \cite{svhn}, and Caltech-101 \cite{caltech}. CIFAR-10 and CIFAR-100 consist of 50,000 training images and 10,000 testing images with the size of 32$\times$32. CIFAR-10 has 10 categories and CIFAR-100 has 100 categories. SVHN consists of 73,257 training images and 26,032 testing images with the size of 32$\times$32. SVHN has 10 classes of digit numbers from `0' to `9'. For training on Cifar-10, Cifar-100, and SVHN, we randomly crop 32$\times$32 images from the 36$\times$36 zero-padded images. Caltech-101 consists of 9,146 images with the size of 300$\times$200. Caltech-101 has 101 semantic categories as well as a background category that there are about 40 to 800 images per category. By following \cite{sraal} we use 90\% of the images for training and 10\% of the images for testing. On Caltech-101, we resize the images to 256$\times$256 and crop 224$\times$224 images at the center. Random horizontal flip and normalization are applied to all the image classification datasets. We summarize the details of the datasets in Table \ref{appendix:dataset}.

\paragraph{Implementation details}

We employ ResNet-18 \cite{resnet} as the image classification model. On all the image classification datasets, the labeling ratio of each active learning cycle is 10\%, 15\%, 20\%, 25\%, 30\%, 35\%, and 40\%, respectively. In an cycle, The model is learned for 200 epochs using an SGD optimizer with a learning rate of 0.1, a momentum of 0.9, a weight decay of 5$\times$10$^{-4}$, and a batch size of 128. After 80\% of the training epochs, the learning rate is decreased to 0.01. We summarize the implementation details in Table \ref{appendix:learning}.

\subsection{Semantic Segmentation}
\paragraph{Dataset}
We evaluate the active learning methods for semantic segmentation on the Cityscapes dataset \cite{cityscapes}. Cityscapes is a large scale driving video dataset collected from urban street scenes of 50 cities. It consists of 2,975 training images and 500 testing images with the size of 2048$\times$1024. By following \cite{vaal}, we convert the dataset from the original 30 classes into 19 classes. We crop 688$\times$688 images from the original images for training. Random horizontal flip and normalization are applied to the images.

\paragraph{Implementation details}

we employ the 22-layer dilated residual network (DRN-D-22) \cite{drn} as the semantic segmentation model. The labeling ratio of each active learning cycle is 10\%, 15\%, 20\%, 25\%, 30\%, 35\%, and 40\%, respectively. In an cycle, the model is learned for 40 epochs using an Adam optimizer \cite{adam} with a learning rate of 5$\times$10$^{-4}$ and a batch size of 4.

\begin{figure*}[t]
    \centering

    \includegraphics[width=0.24\linewidth]{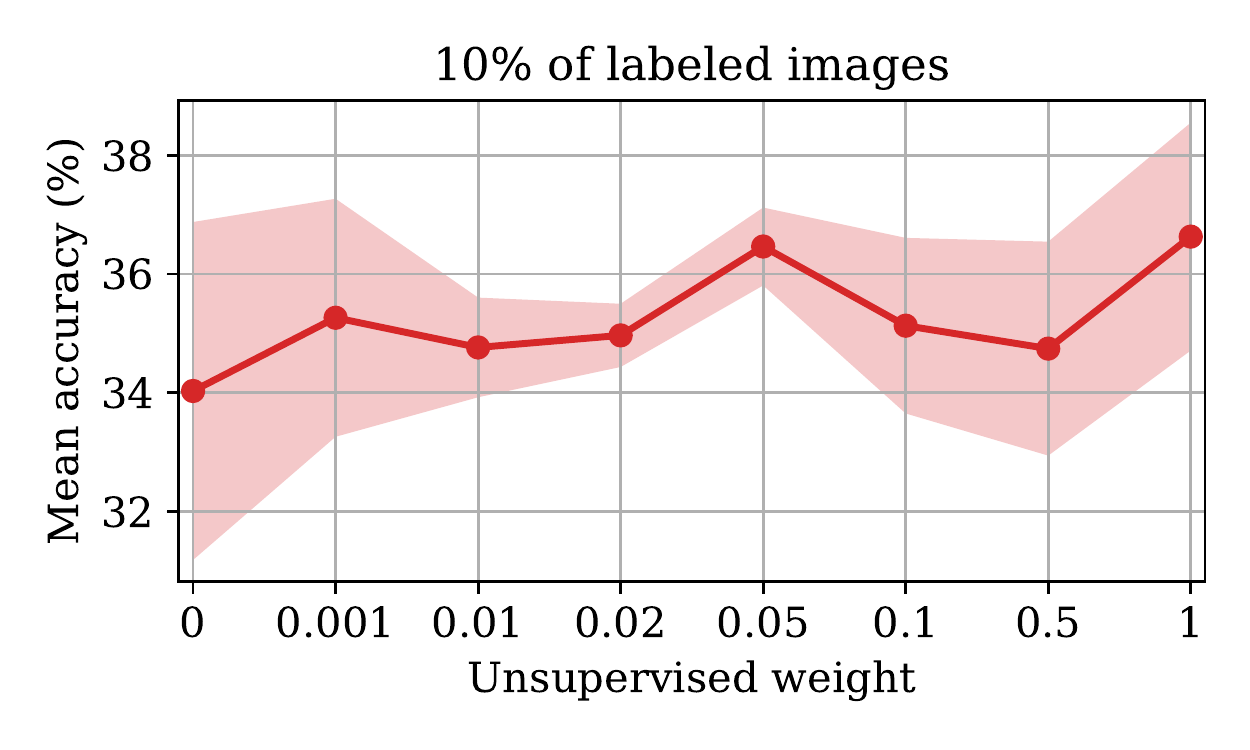}
    \includegraphics[width=0.24\linewidth]{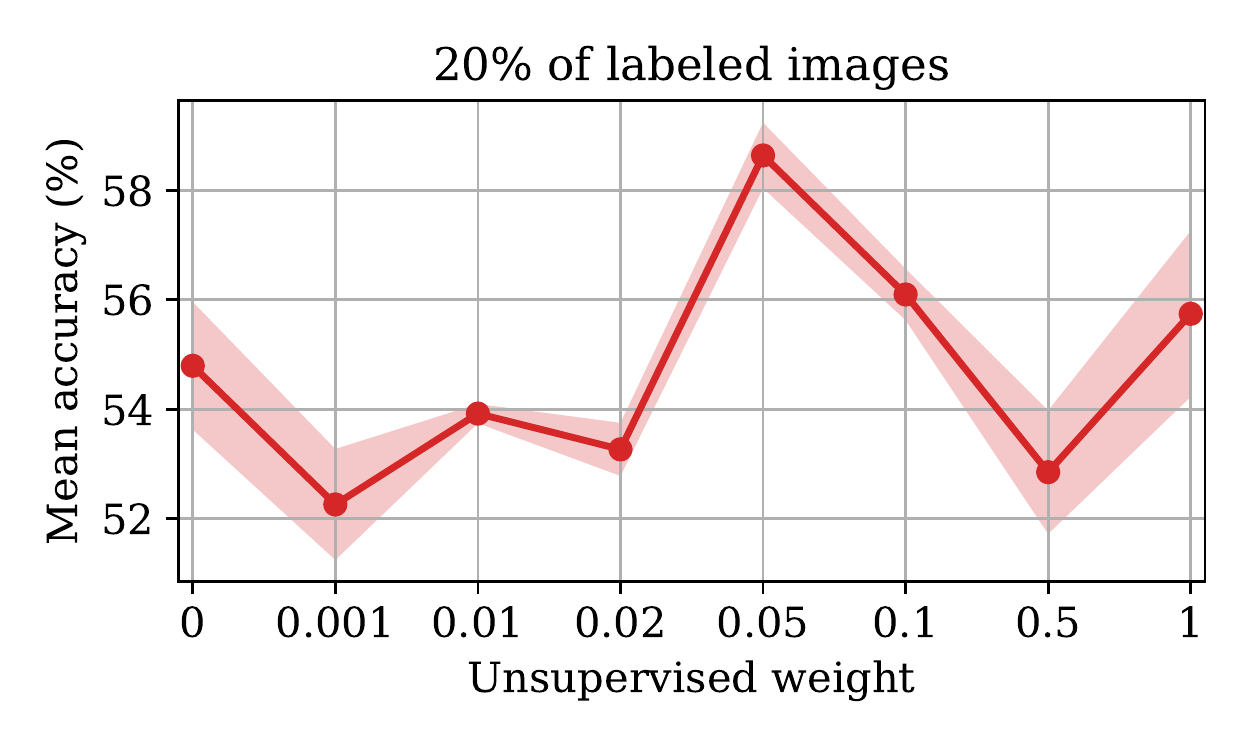}
    \includegraphics[width=0.24\linewidth]{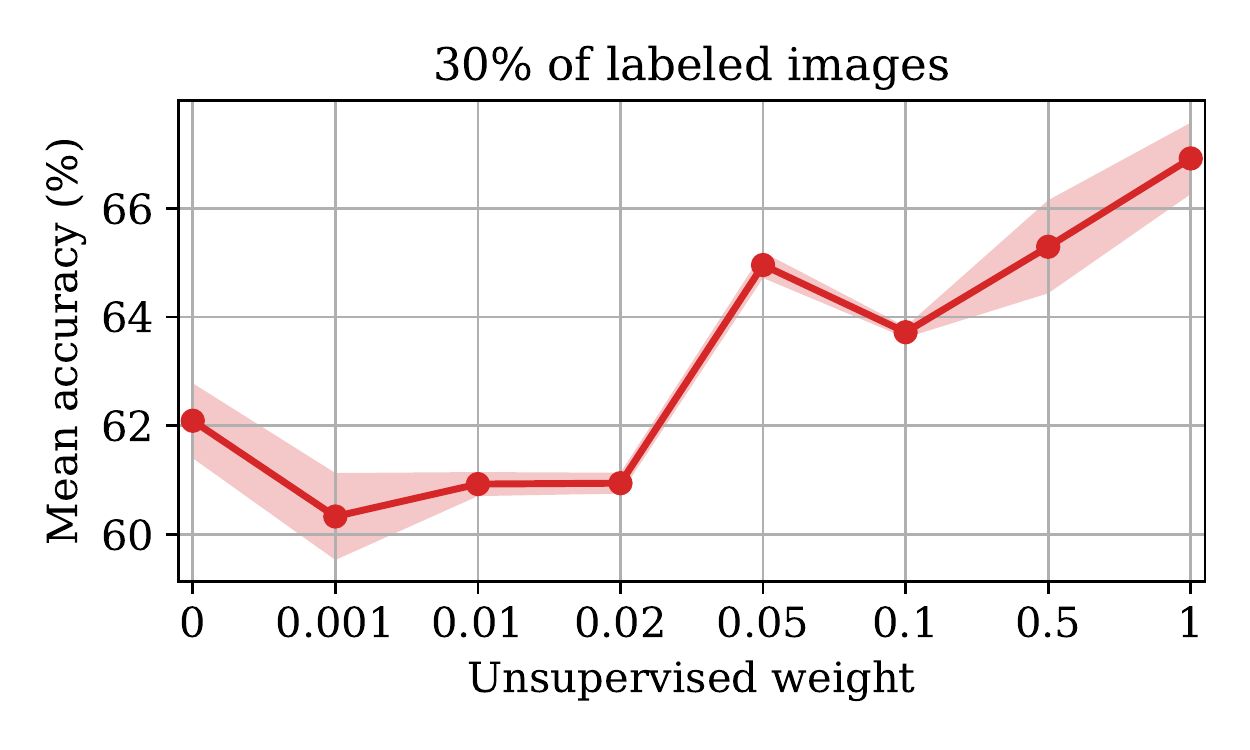}
    \includegraphics[width=0.24\linewidth]{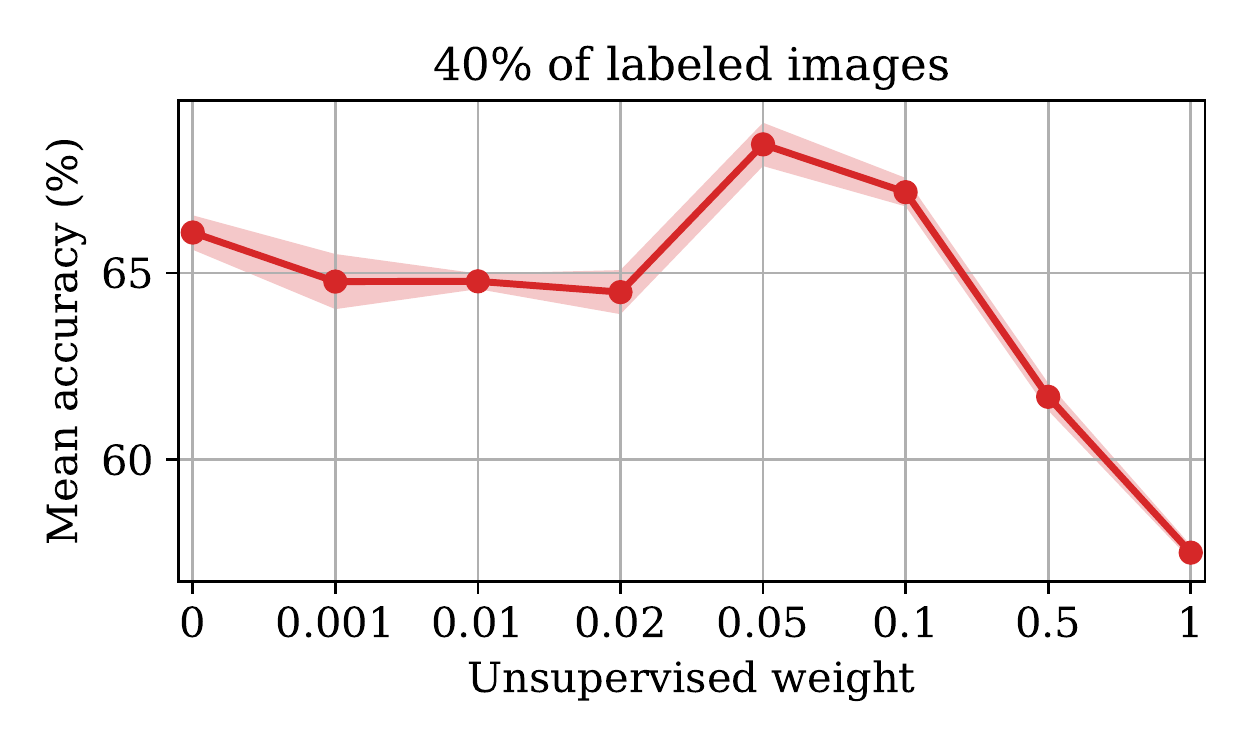}
    \\  
    \includegraphics[width=0.24\linewidth]{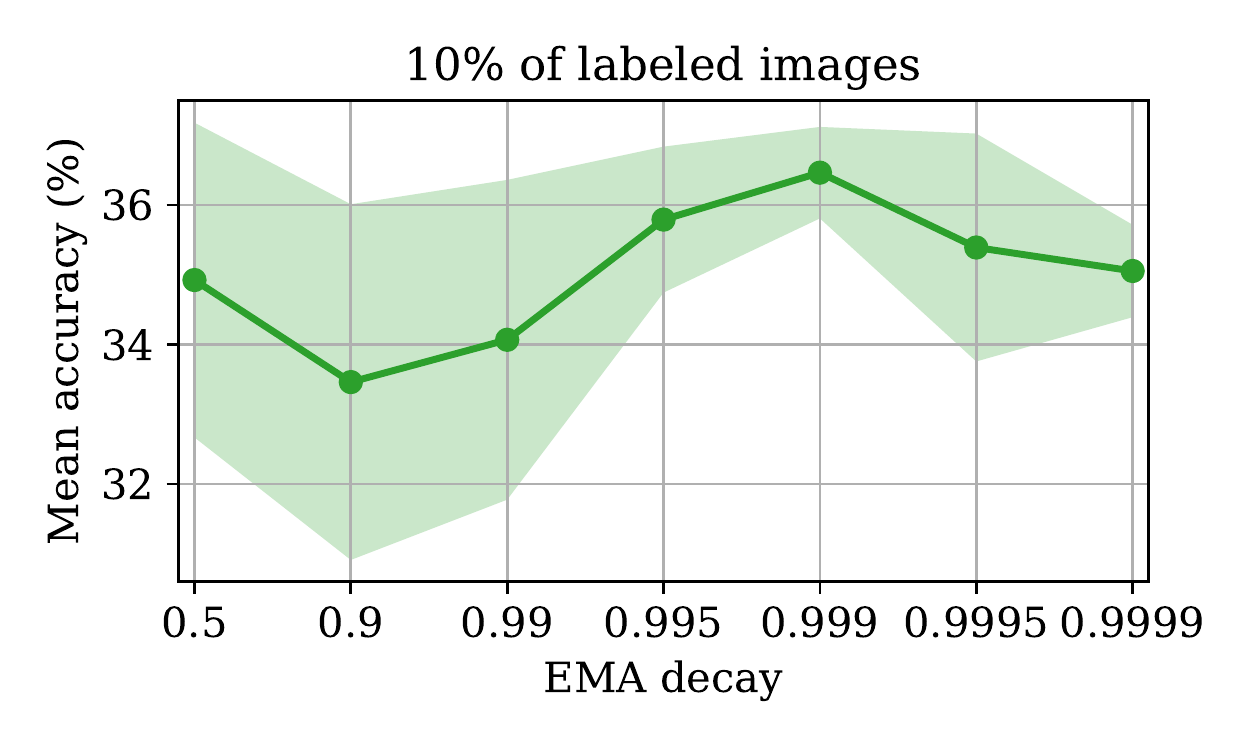}
    \includegraphics[width=0.24\linewidth]{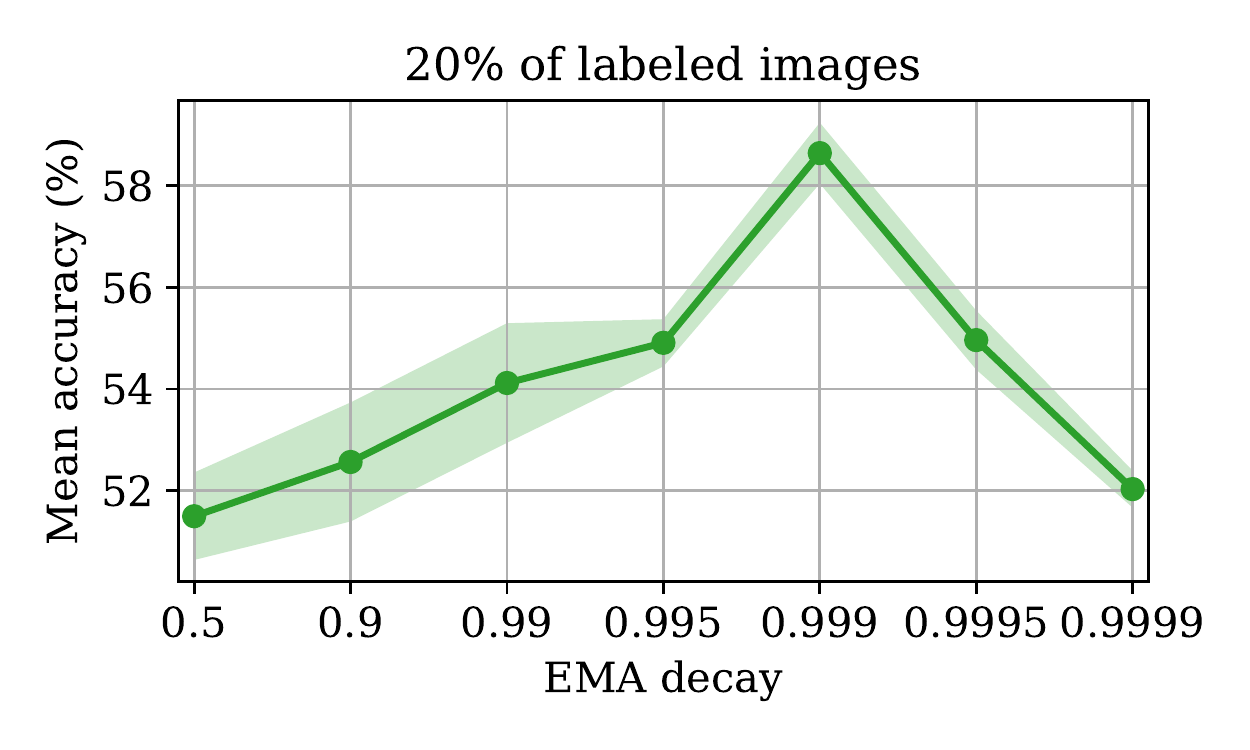}
    \includegraphics[width=0.24\linewidth]{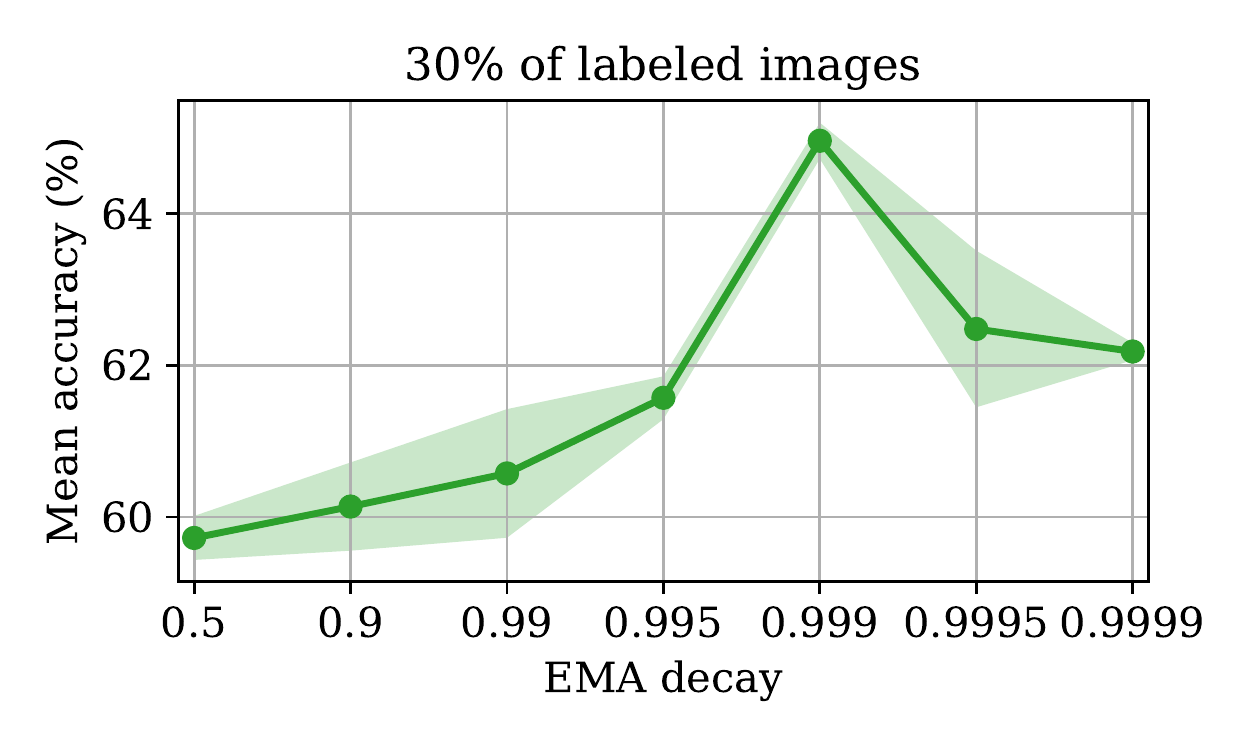}
    \includegraphics[width=0.24\linewidth]{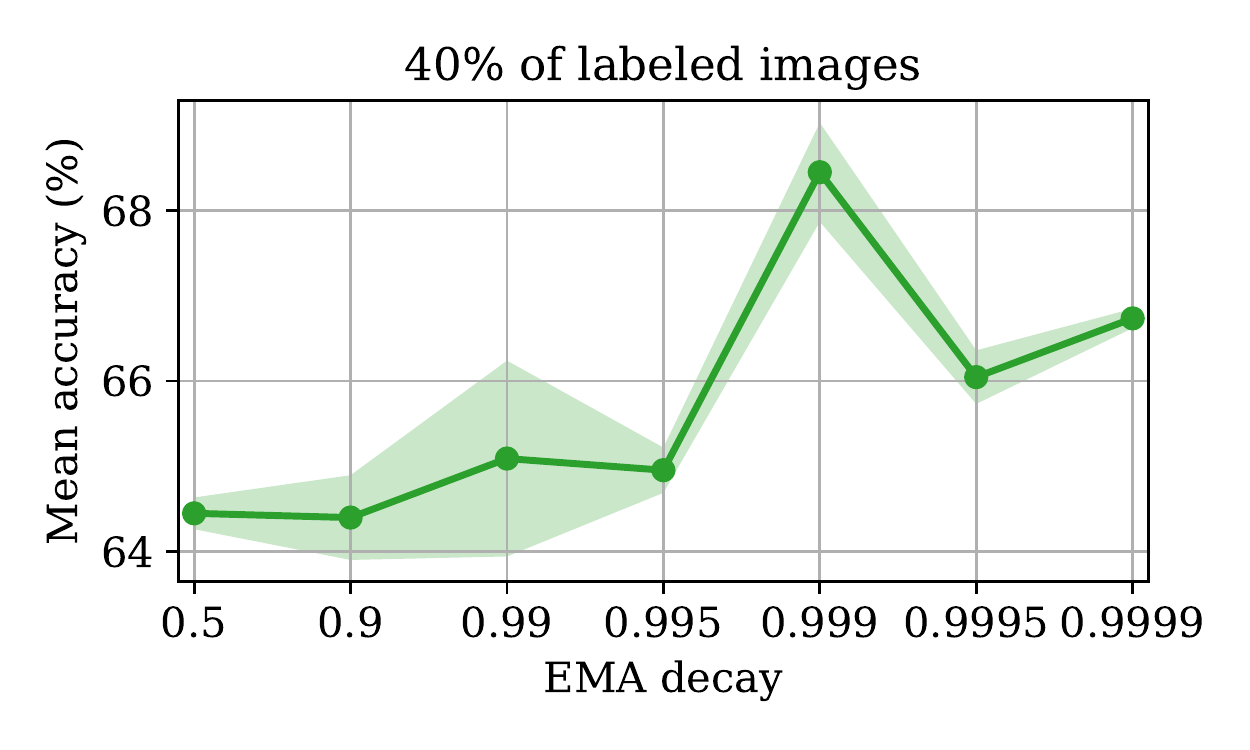}

    % \subfigure[Study on unsupervised loss weight $\lambda$]{
    % \begin{minipage}[b]{1\linewidth}
    % \includegraphics[width=0.246\linewidth]{fig/weight_percent10.pdf}
    % \includegraphics[width=0.246\linewidth]{fig/weight_percent20.pdf}
    % \includegraphics[width=0.246\linewidth]{fig/weight_percent30.pdf}
    % \includegraphics[width=0.246\linewidth]{fig/weight_percent40.pdf}
    % \end{minipage}}
    % \subfigure[Study on EMA decay rate $\alpha$]{
    % \begin{minipage}[b]{1\linewidth}
    % \includegraphics[width=0.246\linewidth]{fig/ema_percent10.pdf}
    % \includegraphics[width=0.246\linewidth]{fig/ema_percent20.pdf}
    % \includegraphics[width=0.246\linewidth]{fig/ema_percent30.pdf}
    % \includegraphics[width=0.246\linewidth]{fig/ema_percent40.pdf}
    % \end{minipage}}
    \caption{Empirical study on unsupervised loss weight $\lambda$ and EMA decay rate $\alpha$. The study is conducted on Cifar-100 with 10\%, 20\%, 30\%, and 40\% of labeled images, respectively. $\lambda=0.05$ and $\alpha=0.999$ achieve the best performances. 
    }

    \label{fig:hyperparameters}
\end{figure*}

\begin{figure*}[h]
    \centering
    \includegraphics[width=0.325\linewidth]{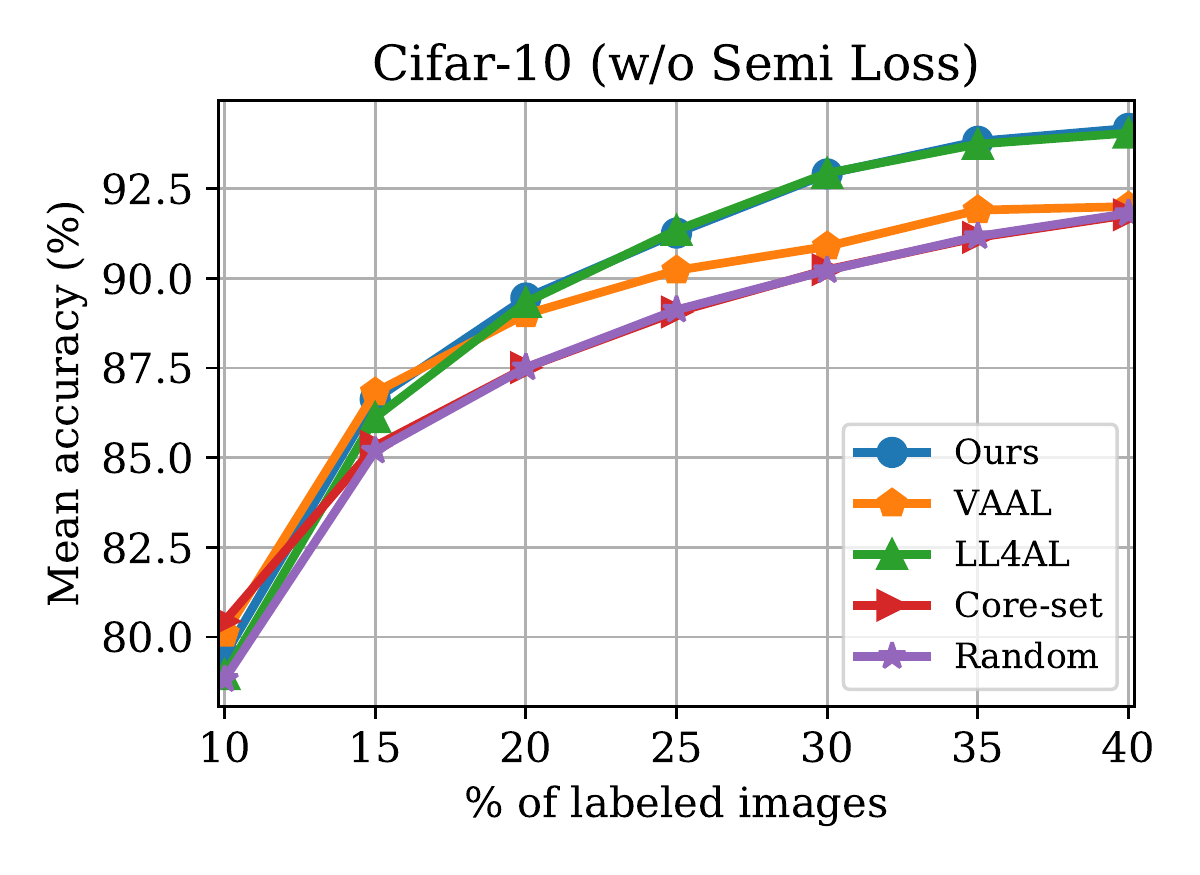}
    \hfill
    \includegraphics[width=0.325\linewidth]{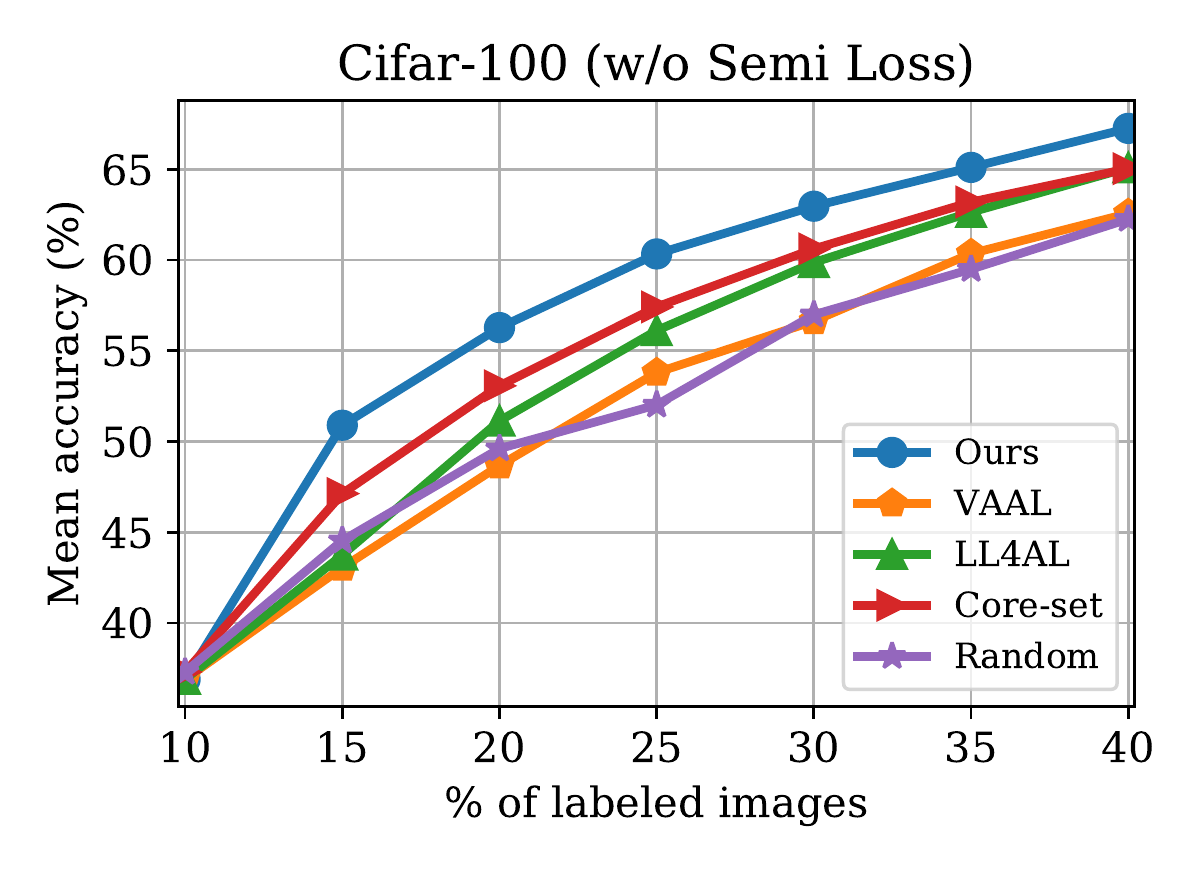}
    \hfill
    \includegraphics[width=0.325\linewidth]{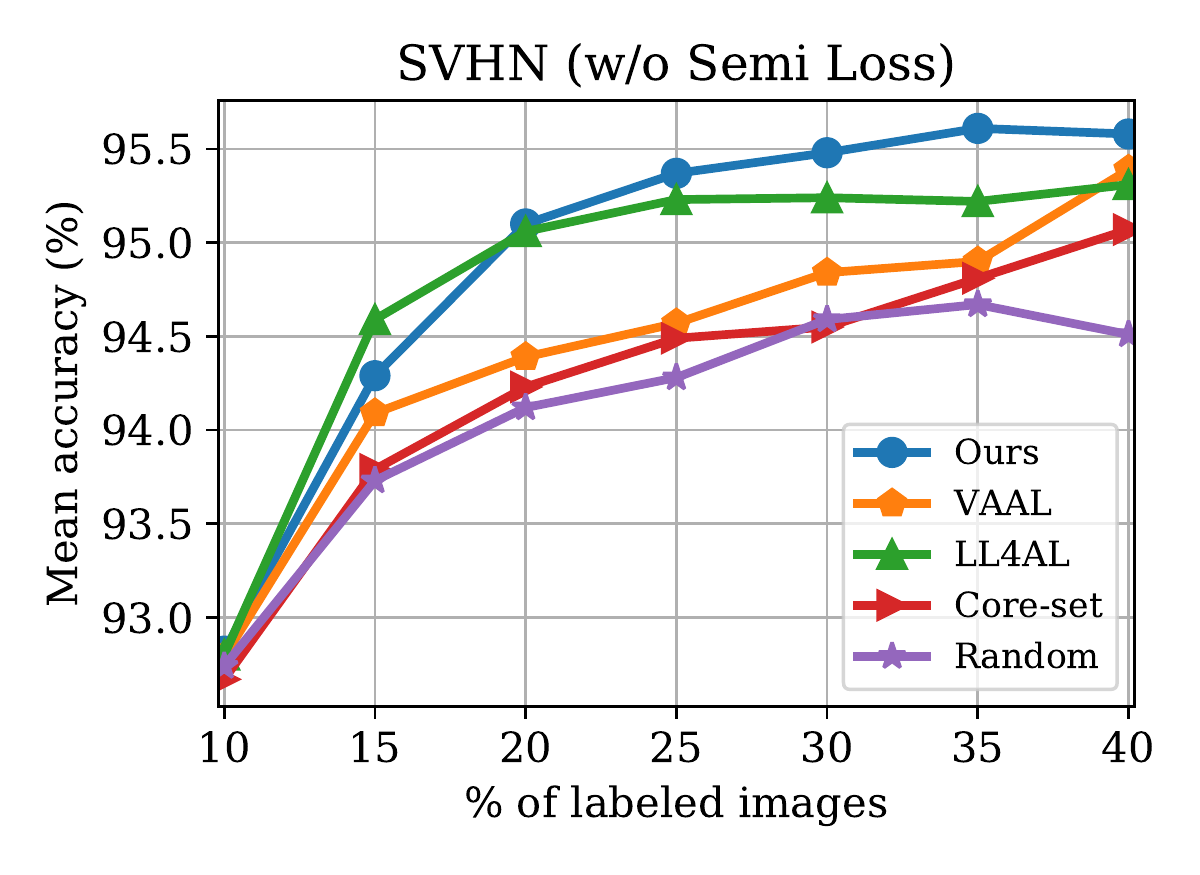} \\
    \includegraphics[width=0.325\linewidth]{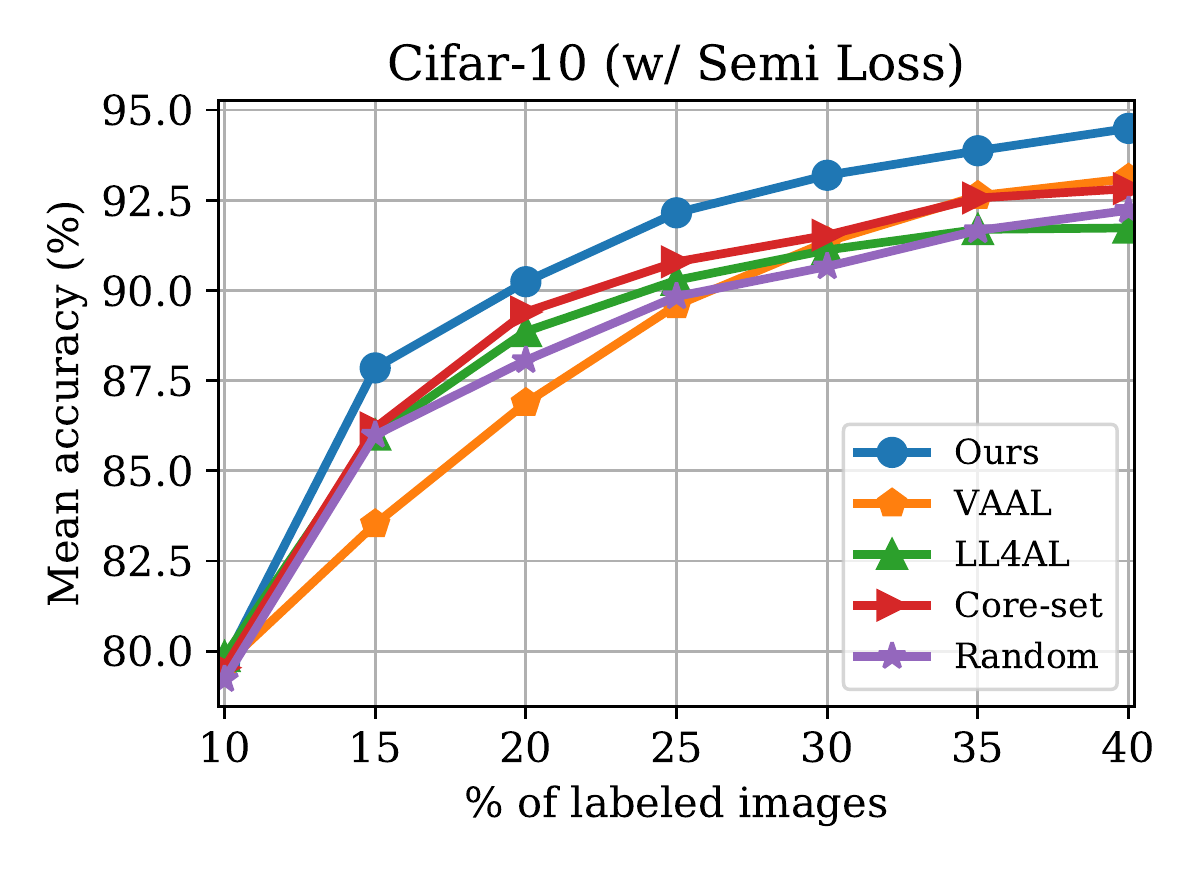}
    \hfill
    \includegraphics[width=0.325\linewidth]{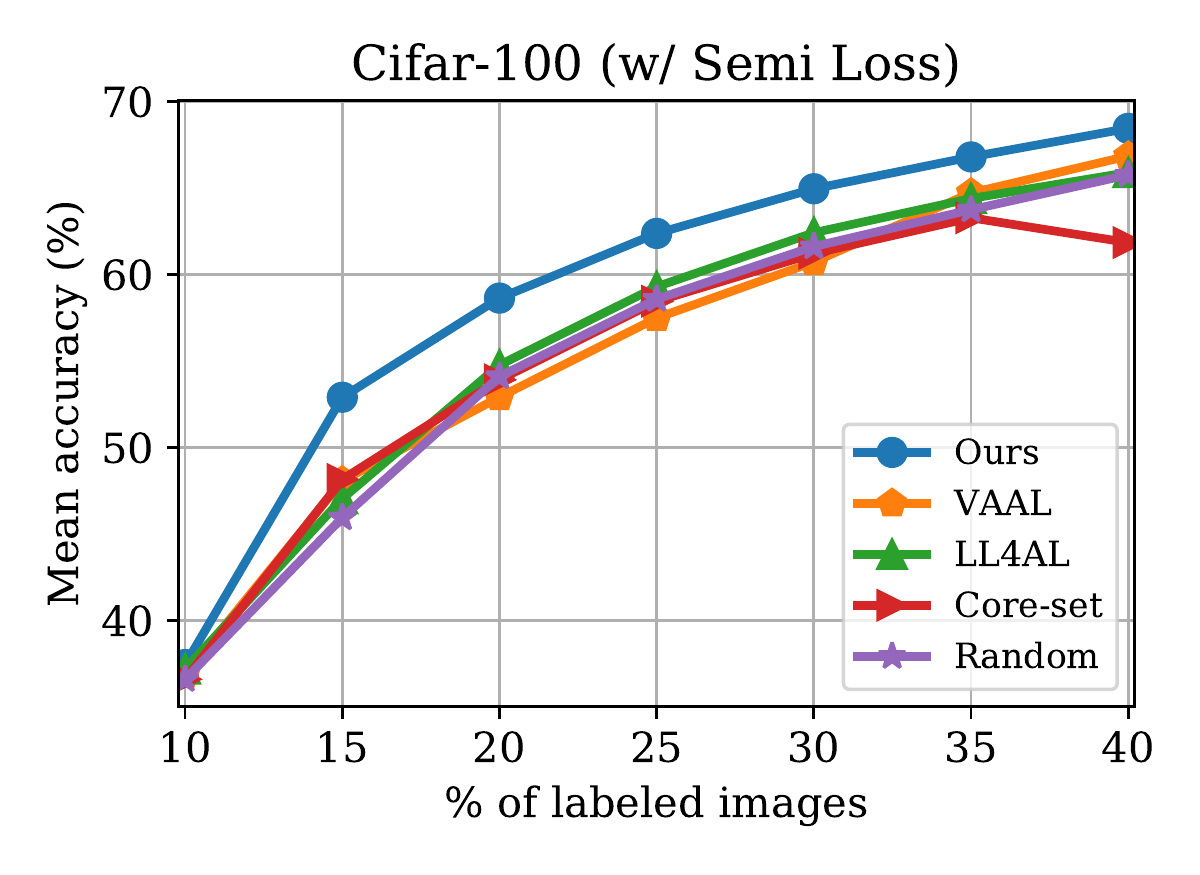}
    \hfill
    \includegraphics[width=0.325\linewidth]{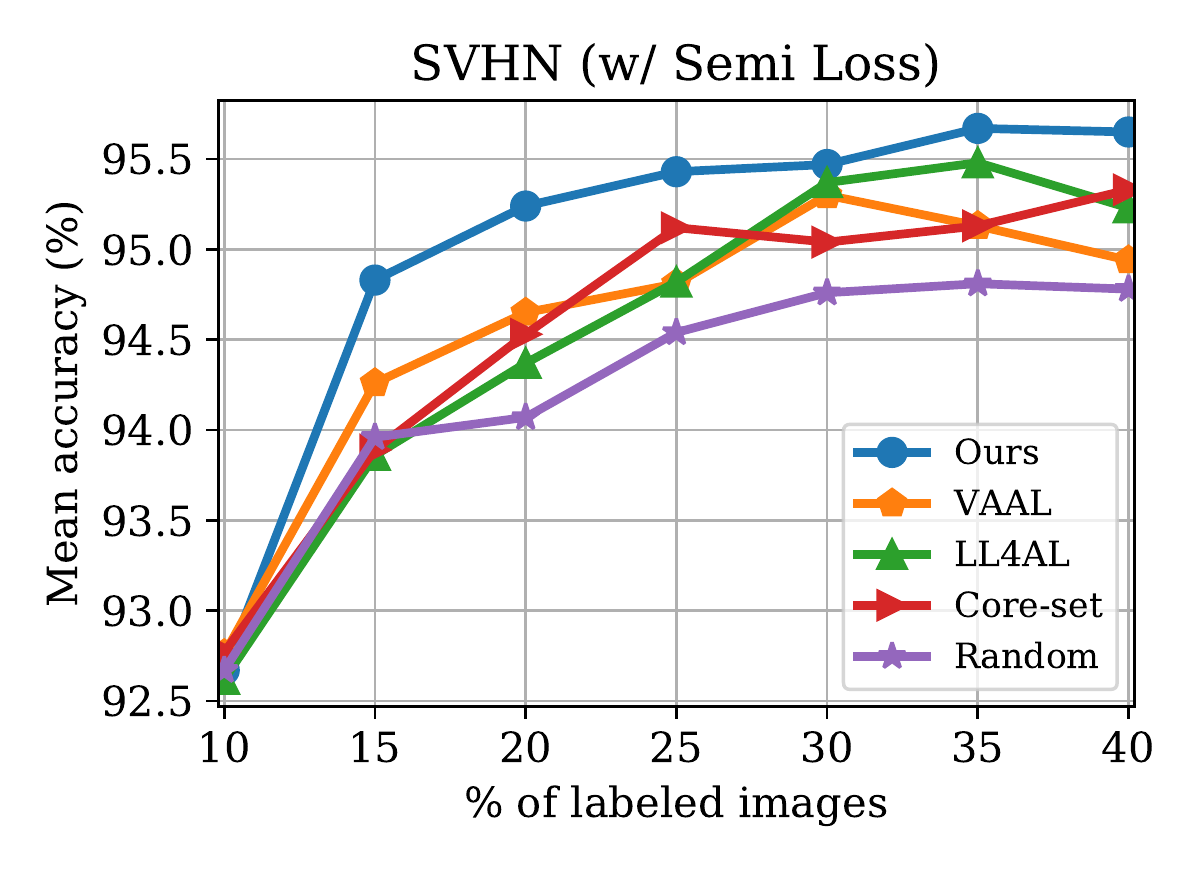} \\
    \caption{The performance of benchmark active learning methods trained \emph{without} (top) and \emph{with} (bottom) the unsupervised loss on three datasets.}
    \label{fig:fig1}
\end{figure*}

\section{More Experimental Results}

\subsection{Study on Hyper-Parameters}
The unsupervised learning plays a vital role in training the task model in active learning. Here we further investigate the hyper-parameter selection for our proposed unsupervised learning method. The hyper-parameters include the unsupervised loss weight $\lambda$ and the EMA decay rate $\alpha$. We conduct empirical studies using our full active learning pipeline on Cifar-100 dataset to investigate the performance variation with different $\lambda$ and $\alpha$. Fig. \ref{fig:hyperparameters} shows the results on labeling budgets of $10\%$, $20\%$, $30\%$, $40\%$, respectively. In most of the cases, $\lambda=0.05$ and $\alpha=0.999$ achieve the best performance. Therefore, we adopt $\lambda=0.05$ and $\alpha=0.999$ for all the experiments in this paper, wherever EMA is involved. 
% When tuning $\lambda$ (top row), the other hyper-parameter $\alpha$ will be fixed, and $\lambda$ will be fixed when tuning $\alpha$ (bottom row). 

\subsection{Evaluating Active Data Selection Strategies}
As a supplementary to Section 5.2 of the paper, we comprehensively compare the active data selection strategies by training the task model with and without the unsupervised loss, respectively. Fig. \ref{fig:fig1} shows that our method achieves superior performances on most of the datasets and settings (either with or without unsupervised loss), demonstrating its effectiveness in active data selection.

\subsection{TOD with Different Gradient Descent Steps}
\begin{figure*}[h]
    \centering
    \includegraphics[width=0.35\linewidth]{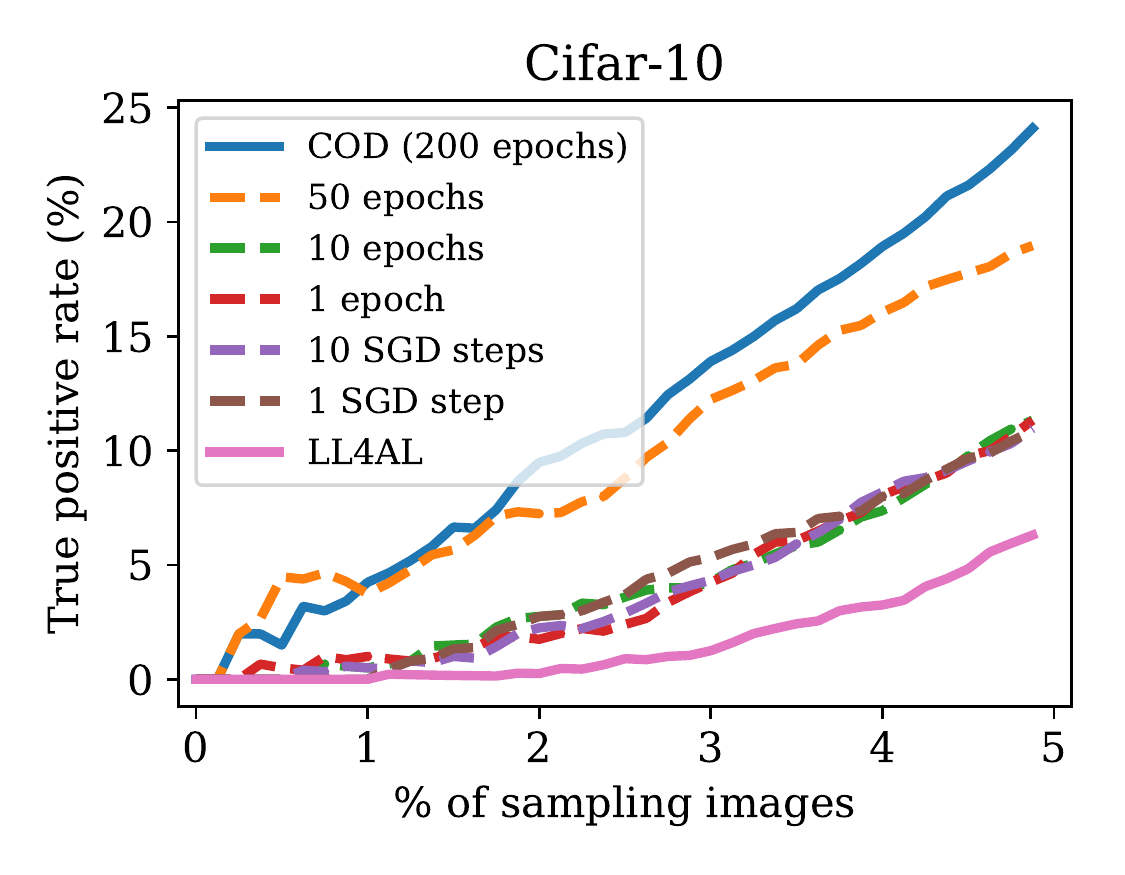}
    \includegraphics[width=0.35\linewidth]{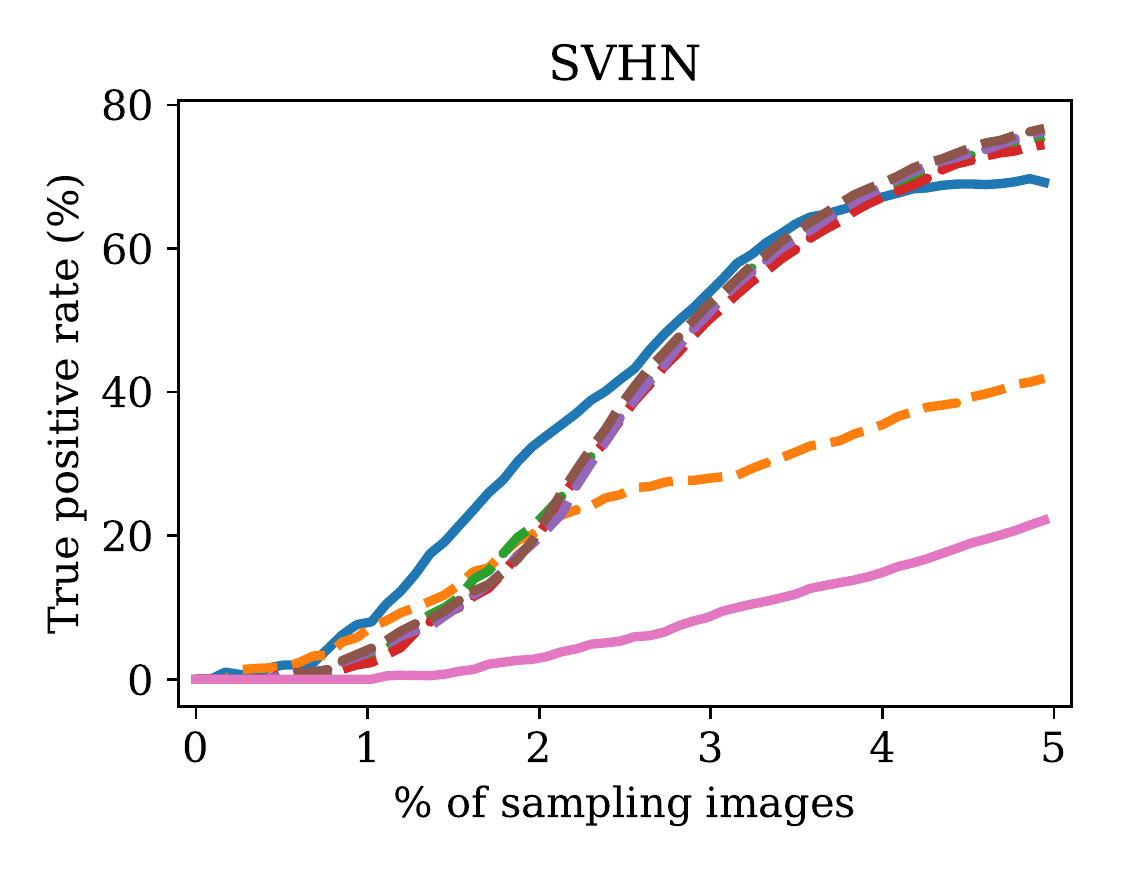} \\
    \caption{The effects of number of GD steps used in TOD.}
    \label{fig:iter_gap}
\end{figure*}

Fig. \ref{fig:iter_gap} shows the loss estimation performances of TOD using different numbers of gradient descent (GD) steps. More GD steps may bring a better loss estimation performance especially when there are fewer sampling images.

\end{document}